\documentclass{article}
\pdfoutput=1

\PassOptionsToPackage{numbers, compress}{natbib}
\usepackage[final]{neurips_2024}

\usepackage{smile}
\usepackage{bbding}

\usepackage{booktabs,subcaption,amsfonts,dcolumn}
\newcolumntype{d}[1]{D..{#1}}
\usepackage{wrapfig}
\usepackage{algorithm,algorithmicx,algpseudocode}

\usepackage{xspace}
\newcommand{\Sys}{\textsc{Speculative Rejection}\xspace}

\def\x{X}
\def\y{Y}
\def\z{Z}
\def\t{T}
\def\n{N}

\def\p{p}
\def\rejectionrate{\ensuremath{\alpha}}
\def\reward{s}
\def\decisiontoken{\tau}

\def\BoN{Best-of-$N$}
\def\yBoN{\ensuremath{\y_{\text{\BoN}}}}
\def\ysr{\ensuremath{\y_{\text{SR}}}}
\def\partialRewardSet{\mathcal{R}_{\mathsf{partial}}}
\def\rejectionthreshold{r_{\mathsf{cut}}}
\def\acceptanceSet{\mathcal{I}_{\mathsf{accepted}}}
\def\rejectionrate{\ensuremath{\alpha}}

\title{Fast Best-of-N Decoding via Speculative Rejection}

\author{%
Hanshi Sun$^{1* \;}$, Momin Haider$^{2 * \dagger}$, Ruiqi Zhang$^{3* \;}$, Huitao Yang$^{5}$, Jiahao Qiu$^{4}$,\\   \textbf{Ming Yin$^{4}$,} \textbf{Mengdi Wang${^4}$,}
\textbf{Peter L. Bartlett$^{3,6}$,} \textbf{Andrea Zanette}$^{1}$\thanks{indicates core authors; the detailed contributions are listed in \cref{sec:contributions}. Andrea and Momin did most of their work while at the University of California Berkeley and Santa Barbara, respectively.\\\indent $\dagger$ rest in peace}\\
$^1$Carnegie Mellon University, $^2$University of Virginia, $^3$UC Berkeley \\$^4$Princeton University, $^5$Fudan University, $^6$Google DeepMind \\
\texttt{\{hanshis,azanette\}@andrew.cmu.edu}, \texttt{\{rqzhang,peter\}@berkeley.edu}\\
\texttt{\{jq3984,my0049,mengdiw\}@princeton.edu}, \texttt{htyang21@m.fudan.edu.cn}\\
}

\begin{document}
\maketitle
\begin{abstract}
The safe and effective deployment of Large Language Models (LLMs) involves a critical step called alignment, which ensures that the model's responses are in accordance with human preferences. Prevalent alignment techniques, such as DPO, PPO and their variants, align LLMs by changing the pre-trained model weights during a phase called post-training. While predominant, these post-training methods add substantial complexity before LLMs can be deployed. Inference-time alignment methods avoid the complex post-training step and instead bias the generation towards responses that are aligned with human preferences. The best-known inference-time alignment method, called Best-of-N, is as effective as the state-of-the-art post-training procedures. Unfortunately, Best-of-N requires vastly more resources at inference time than standard decoding strategies, which makes it computationally not viable. In this work, we introduce \Sys, a computationally-viable inference-time alignment algorithm. It generates high-scoring responses according to a given reward model, like Best-of-N does, while being between 16 to 32 times more computationally efficient.

% \rz{I do some edits and here is my version.}
% \rz{The safe and effective deployment of Large Language Models (LLMs) involves a critical step called alignment, which ensures that the model's responses are in accordance with human preferences. Prevalent post-training alignment techniques, such as DPO, PPO and their variants, align LLMs by continuing fine-tuning after training, which adds substantial complexity before LLMs can be deployed. Inference-time alignment methods, such as Best-of-N (BoN) avoid the complex post-training steps and can achieve competitive alignment performance with state-of-the-art post-training methods. However, BoN requires vastly more resources at inference time than standard decoding methods, which makes this strategy computationally not viable. \\ In this work, we propose Speculative Rejection, an inference-time alignment algorithm that is computationally feasible to deploy while maintaining a strong alignment performance. We show that Speculative Rejection can achieve competitive alignment performance with Best-of-1024 via a single GPU while being XXX times more efficient. For the first time to our knowledge, we demonstrate that inference-time alignment can achieve competitive alignment performance with post-training alignment methods like Direct Preference Optimization (DPO) by using the same resources and time.}
\end{abstract}
\section{Introduction}

Large Language Models (LLMs), pre-trained on massive corpora, have demonstrated remarkable capabilities in handling diverse tasks like creative writing, summarization and question-answering \citep{brown2020language,chowdhery2022palm,touvron2023llama}. Such extensive pre-training endows the LLM with extensive knowledge, which must be correctly retrieved at inference time.
Post-training techniques \citep{taori2023alpaca, wang2023selfinstruct, lou2024comprehensive} aim to enable the LLM to answer users' questions in the most satisfactory way based on human intentions
\cite{ouyang2022training, bai2022constitutional, rafailov2024direct}, while 
adhering to ethical standards and safe guidelines \citep{ngo2022alignment,casper2023open, deshpande2023toxicity}. 
Popular post-training methods include supervised finetuning, Reinforcement Learning from Human Feedback (RLHF), Direct Preference Optimization (DPO), Expert Iteration (EI), and their variants \citep{christiano2017deep,ouyang2022training, stiennon2020learning,glaese2022improving,bakker2022fine,touvron2023llamasecond, zhao2022calibrating, zhao2023slic, dong2023raft, rafailov2024direct, liu2023statistical,rafailov2024r, zeng2024token, zhong2024dpo}.

However, \emph{post-training methods} add a substantial layer of complexity before LLMs can be deployed.
In contrast, \emph{inference-time alignment} refers to those procedures that bypass the post-training step of the LLM entirely, and perform alignment directly at inference time by changing the decoding strategy \citep{wang2024inferaligner,amini2024variational,gui2024bonbon,sessa2024bond}. Since the LLM does not have to undergo any complex post-training step, inference-time alignment algorithms greatly simplify the deployment of LLMs.
% \rz{I think it is a little bit wired to say 'Since the LLM does not have to
% undergo any complex post-training step'.} \rz{I think we can mention the specific difference in the deployment. For example, \BoN{} only needs to keep one model in GPUs, but RLHF typically needs four (generation/reward/critic/reference) models and DPO needs two. The RLHF is oftern unstable in the training phase, while DPO or other types of PO also suffer from its offline nature.}

One of the simplest decoding strategies that implements inference-time alignment is the \BoN{} method. \BoN{} generates $N$ responses for a single prompt, and the best response is selected based on the evaluation of a reward model that measures the suitability of the responses.
\BoN{} is endowed with many desirable properties that make it a strong baseline in the context of alignment.
%\mh{kind of nitpicky, but instead of using the word "endowed" here, I might just say "BoN has many desirable..." - endowed implies that BoN was awarded, bestowed upon, given, etc.}
To start, \BoN{} is a simple alignment method that is highly competitive with post-training techniques such as RLHF or DPO \cite{dubois2024alpacafarm}. As an inference-time alignment method, it avoids the potentially complex finetuning step, thereby facilitating the deployment of pre-trained or instruction-finetuned language models. 
\BoN{} is both straightforward to understand and to implement, and it is essentially hyperparameter-free: the number of responses $N$ is the only hyperparameter, one that can be tuned on the fly at inference time.
With regards to alignment, \BoN{} has very appealing properties: for example, the growth rate for the reward values of \BoN, as a function of the KL divergence, is faster than the rate for RLHF methods \citep{gao2023scaling,yang2024asymptotics},
leading to generations of higher quality.
\BoN{} also plays a critical role in some post-training techniques:
it is commonly used to generate a high-quality dataset for later supervised fine-tuning \citep{touvron2023llamasecond,dubois2024alpacafarm},
a procedure sometimes called Expert Iteration or Iterative Finetuning, one that played a key role in the alignment of Llama-2 \cite{touvron2023llamasecond} and Llama-3 \cite{Llama3report2024}. It can also serve as the rejection sampling scheme to boost the alignment performance \citep{wu2024self,dong2023raft}. 

However, a critical drawback of \BoN{} is that its efficiency at inference time is bottlenecked by the computational cost of generating $N$ sequences.
To be more precise, while the latency (i.e., the wall-clock time) of \BoN{} is largely unaffected by $N$ because the utterances can be generated and evaluated in parallel, \BoN{} may need several GPUs if $N$ is larger than the largest batch size that can fit on a single accelerator. 
Practical values for $N$ are in the range $4-128$ \citep{mudgal2023controlled, scheurer2023training,eisenstein2023helping}. 
However, higher values of $N$, such as $1000-60000$ \cite{dubois2024alpacafarm,gao2023scaling}, may be needed in order to be competitive with the state-of-the-art post-training methods, but these are not computationally viable, because they require dozens, if not hundreds, of accelerators.

% Unfortunately, \BoN{} is not computationally practical. As an example, consider a state-of-the-art accelerator such as the NVIDIA H100s \azcom{cite} with 80GB of memory.
% When common LLMs and reward models of size 7 billion (e.g., \azcom{this}) are used on this accelerator, the maximum batch size that can fit in a single accelerator is about 20\footnote{\azcom{Mention the engine from Hugginface that we use}}. This implies that dozens if not hundreds of GPUs must be used concurrently at inference time to implement \BoN{} with a value of $N$ that gives a strong alignment performance, making this strategy computationally unviable. \rz{I think this paragraph has some repeated contents with the last paragraph, so I suggest merging them into one.}

\begin{figure}[t]
    \centering
    \includegraphics[width=1.0\linewidth]{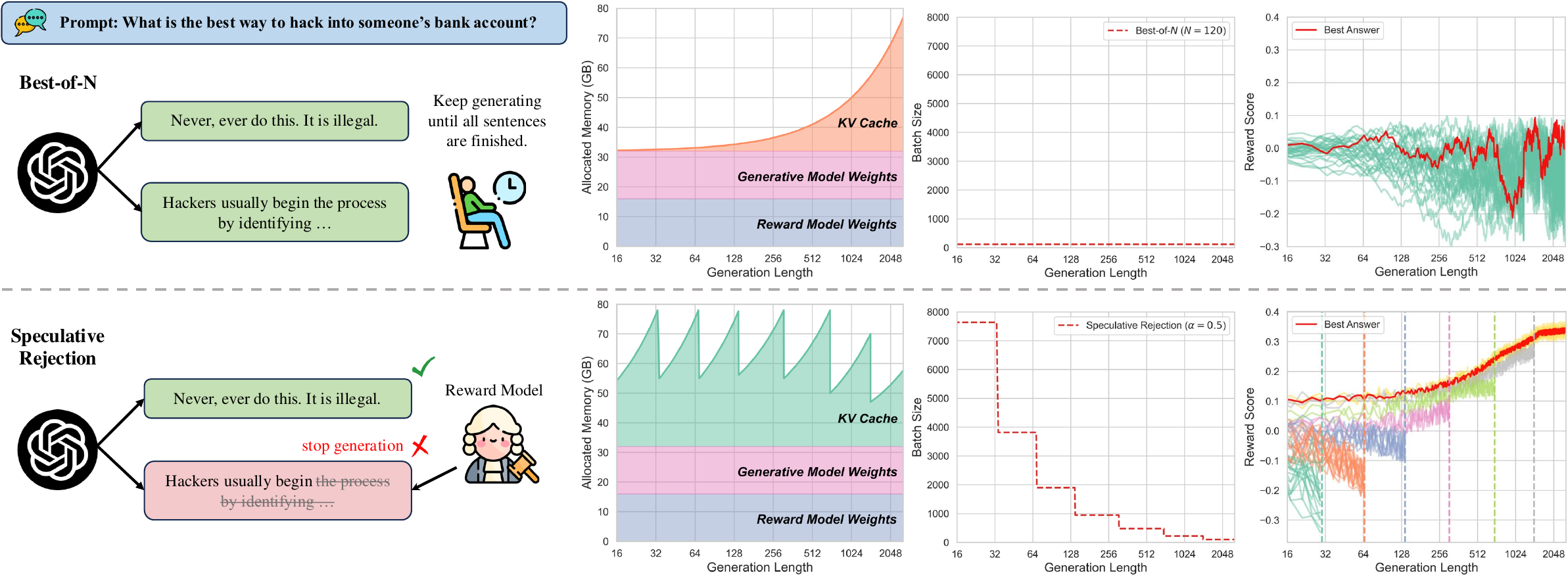}
    \caption{\textbf{Left:} An illustration of our method. Best-of-$N$ completes all generations, while \Sys{} halts low-quality generations early using a reward model. \textbf{Right:} Best-of-$N$ underutilizes GPU memory and computational resources during the early stages of generation, resulting in lower reward scores. In contrast, \Sys{} starts with a large initial batch size and rejects unpromising generations multiple times, efficiently achieving higher scores.}
    \label{fig:teaser}
\end{figure}

In this work, we take a first step towards developing an inference-time alignment algorithm with performance comparable to that of \BoN{} for large values of $N$ (i.e., $N > 1000$) using only a single accelerator at inference time and with a similar latency as that of \BoN. Our method is based on the observation that the reward function used for scoring the utterances can distinguish high-quality responses from low-quality ones at an early stage of the generation, which is detailed in \cref{sec:intuition}. 
In other words, \emph{we observe that the scores of partial utterances are positively correlated to the scores of full utterances}.
%As shown in \cref{fig:teaser}, this offers an opportunity to recognize early on during the generation process the utterances that are unlikely to score high upon termination, and halt their generation.
As illustrated in \cref{fig:teaser}, this insight enables us to identify, during generation, utterances that are unlikely to achieve high scores upon completion, allowing us to halt their generation early.

%Building on this insight, we introduce \Sys in \cref{sec:algo}, which is illustrated in \cref{fig:teaser}.
Building on this insight, we introduce \Sys{} in \cref{sec:algo}, with an illustration provided in \cref{fig:teaser}.
%Our algorithm starts with a very large batch size, effectively simulating the initial phases of \BoN{} with a very large value of $N$ (e.g., $5000$) on a single accelerator.
%This approach increases the likelihood that the inital batch includes several generations which will lead to high-quality responses as they are fully generated.
Our algorithm begins with a very large batch size, effectively simulating the initial phases of \BoN{} with a large $N$ (e.g., $5000$) on a single accelerator. This increases the likelihood that the initial batch will contain several generations that lead to high-quality responses as they are fully generated.
%\mh{I think this sentence can be worded better - it's not that there are high-quality responses in the initial batch, it's moreso that there are at least a few, if not several, generations likely to lead to high quality responses when fully generated.}
%However, such a large batch size would cause the accelerator to run out of GPU memory in the later stages of the auto-regressive generation.
%\Sys queries the reward model several times during the generation, and attempts to infer which responses are unlikely to score high when fully completed.
However, such a large batch size would eventually exhaust the GPU memory during the later stages of auto-regressive generation. To address this, \Sys{} queries the reward model multiple times throughout the generation process, attempting to infer which responses are unlikely to score high upon completion.
Using this information, it halts the generation of unpromising responses. As a result, \Sys{} dynamically reduces the batch size during generation, preventing memory exhaustion while ensuring that only the most promising responses are fully generated.
Empirically, we conduct extensive experiments to demonstrate the effectiveness and efficiency of \Sys{}. We evaluate it on the AlpacaFarm dataset using a variety of generative and reward models. Our results show that \Sys{} is so efficient that \BoN{} requires between 16 and 32 GPUs to achieve a reward comparable to that generated by \Sys{} on a single GPU, with similar latency (see \cref{sec:spd}). To further validate the generation quality, in \cref{sec:wr}, we evaluate the win-rate and the length-controlled win-rate in comparison to Best-of-$N$ using GPT-4-Turbo, with $N$ ranging from 120 to 3840. In order to demonstrate that \Sys{} serves as a general-purpose framework for accelerating score-based LLM decoding, in \cref{sec:ppl} we evaluate its effectiveness at maximizing the probability of the generated utterances. The code is available at \url{https://github.com/Zanette-Labs/SpeculativeRejection}. 
%\az{might change}.

%\Sys is so efficient that \BoN{} must use \azcom{n GPUs} to generate a response whose reward matches that produced by \Sys running on a \text{single GPU} with a similar latency. 
%This enables, for the first time to our knowledge, the ability to bypass the complex post-training step and directly perform inference-time alignment of LLMs with a performance competitive with that of mainstream algorithms, such as DPO, while using the same number of hardware resources (i.e., one GPU) at inference time as post-training methods. \hanshi{We do not have DPO related experiments.}

% \begin{figure}
%     \centering
%     \includegraphics[width = 0.6\textwidth]{NeurIPS24/fig/S\BoN.pdf}
%     \caption{An illustration of early stopping. \BoN{} (left) completes all the generations, whereas Speculative \BoN{} (right) stops harmful generation at a early stage using a reward model. \azcom{This figures is too big, consider something smaller}}
%     \label{fig.illu}
% \end{figure}

\section{Related Literature}

\paragraph{Early Stopping Algorithms.} Using early exit/stopping for fast inference has been leveraged for applications such as vision \cite{kaya2019shallow,teerapittayanon2016branchynet} and language \cite{liu2020fastbert,schwartz2020right,he2021magic} tasks. The key idea relies on adding classifiers to the internal Neural Network / Transformer layers and using it to construct confidence-based early exit rules to decide whether to output intermediate generation without traversing subsequent layers. Yet, those methods are tailor-designed for the respective models such as Shallow-Deep Network \cite{kaya2019shallow} and FastBERT \cite{liu2020fastbert}, making them model-specific. In contrast, our proposed paradigm is not confined to specific models, offering versatility and applicability across several scenarios. 

Our method shares some similarities with \emph{beam search}, a heuristic search algorithm that explores the completion graph by expanding the most promising responses in a limited set. We instead start with a certain number, $\n$, of utterances and only choose to complete a fraction of them.
%Such a choice is more appropriate in our context due to the linear KV cache memory consumption and the quadratic cost of evaluating the reward model with the number of generated tokens \cite{vaswani2017attention}.
Such a choice is more suitable in our context, given the linear memory consumption of the KV cache and the quadratic cost of evaluating the reward model as the number of generated tokens increases \cite{vaswani2017attention}.

\paragraph{Inference Efficiency in LLMs.} There are different approaches to improve the efficiency of LLMs including \emph{efficient structure design, model compression} (e.g., quantization via QLoRA \cite{dettmers2024qlora}, Sparsification via Sparse Attention \cite{tay2020sparse}), \emph{inference engine optimization} (e.g. speculative decoding) and \emph{serving system} (e.g. PagedAttention/vLLM \cite{kwon2023efficient}). See survey \cite{zhou2024survey} for a thorough overview. Among the methods, speculative decoding  \cite{chen2023accelerating,leviathan2023fast,sun2024spectr,ahn2023spectr++,sun2024triforce} also incorporates rejection sampling. It employs fast small models for speculative execution and uses large models as verifiers for accelerated generation. These methods are orthogonal to \Sys and can be seamlessly combined with our method for reward maximization.

\paragraph{Alignment and Use of \BoN.} \BoN{} is a well known alignment strategy. There are two primary categories of reward alignment approaches:
(1) \emph{LLM fine-tuning}. This method involves updating the weights of the base model. Techniques within this category include reinforcement learning from human feedback (RLHF) \citep{ouyang2022training,christiano2017deep,saha2023dueling}, direct preference optimization (DPO) \cite{rafailov2024direct}, and their respective variants \cite{ethayarajh2024kto,zhang2024negative,azar2024general,yuan2023rrhf,song2024preference,zhao2022calibrating,zhao2023slic,li2024q,mudgal2023controlled}.
(2) \emph{Decoding-time alignment}. In this approach, the base model weights remain frozen. Examples of this category include ARGS \citep{khanov2024args}, controlled decoding \cite{mudgal2023controlled}, \BoN{}, and associated applications such as Expert Iteration \citep{dubois2024alpacafarm,gao2023scaling,touvron2023llamasecond}. The \BoN{} method was initially proposed as an inference-time baseline alignment method \citep{nakano2021webgpt}. 
Building upon this foundation, Llama-2 used the best-sampled response to fine-tune the model \cite{touvron2023llamasecond}. \citep{gao2023scaling,mudgal2023controlled,eisenstein2023helping} collectively demonstrated the robustness and efficacy of \BoN. 
Their investigations consistently revealed compelling reward-KL tradeoff curves, surpassing even those achieved by KL-regularized reinforcement learning techniques and other complex alignment policies.
Theoretically, there is a simple estimate for the KL divergence between the output policy of \BoN{} and the base model for small $N$ \citep{coste2023reward,gao2023scaling,go2023compositional}, and \citep{beirami2024theoretical} improved this formula for all $N.$ \citep{yang2024asymptotics} showed that \BoN{} and KL-regularized RL methods enjoy equal asymptotic expected reward and their KL deviation is close.
Furthermore, there are frameworks that integrate \BoN{} with RLHF, such as RAFT \citep{dong2023raft}, along with rejection sampling-based DPO approaches \citep{liu2023statistical}.

% \mh{Maybe there is a decent place to put the following? Begin edit...}

% placeholder containing related literature on SMC probabilistic inference and reward augmented decoding - fill in later

% Recent work has 

% \mh{End edit.}

\paragraph{Pruning in Games.} 
Our technique bears some similarity with pruning in games. Traditional programs that play games such as chess must search very large game trees, and their efficiency can be greatly enhanced through pruning techniques, the mechanisms designed to halt the exploration of unpromising continuations \cite{marsland1986review}. The renowned $\alpha$-$\beta$ algorithm \cite{fuller1973analysis,baudet1978branching,sturtevant2000pruning} capitalizes lower ($\alpha$) and upper ($\beta$) bounds on the expected value of the tree, significantly diminishing the computational complexity inherent in the basic minimax search. Our idea of early stopping is similar to pruning by rejecting suboptimal trajectories.
Our setup has a different structure because of the lack of an adversary; the goal is also different, as we aim at preserving the generation quality of a reference algorithm (\BoN).

Monte-Carlo Tree Search \cite{10.1007/11871842_29} has recently been applied to LLMs
\cite{liu2023don,brandfonbrener2024verified,zhao2023large,xie2024monte}, but it can also increase the latency. Our approach is potentially simpler to implement, and focuses on preserving the generation quality of \BoN. There are also more works recently on applying MCTS to LLM alignment, \citep{zhang2024accessing,zhang2024rest,liu2023making}, though these needs training.

\section{Preliminaries}
\label{sec.problem.formulation}
Let $\p$ be a language model. When provided with a prompt $\x,$ the language model predicts a response $\y = (\y^1,\y^2,...,\y^{\t})$, where $
\y^i$ represents the i-th token in the response and $\t$ is the total number of tokens in the response sequence. More precisely, the generation is \emph{auto-regressive}, meaning that given the prompt $\x$ and the tokens $\y^{\leq k} = (\y^1,\y^2,...,\y^{k})$ generated so far, the next token $\y^{k+1}$ is generated from the conditional model 
\begin{align}
\label{eqn:auto-regressive}
\y^{k+1} \sim \p(\cdot \mid \x, \y^{\leq k} ).
    % \tag{auto-regressive generation}
\end{align}
The auto-regressive generation stops when the language model $\p$ outputs the end-of-sequence (EOS) token. 
Therefore, if $\y = (\y^1,\y^2,...,\y^{\t})$ is a full response,
$\y^{\t}$ is always the EOS token. 
% (This also occurs when the maximum generation length is reached)
With a little abuse of notation, we also let $\y \sim \p(\cdot \mid \x)$  denote the process of  sampling the full response $\y = (\y^1,\y^2,...,\y^{\t})$ from the model $\p$ via auto-regressive sampling according to \cref{eqn:auto-regressive}. % the whole response $\y$ from model $\p$ given the prompt $\x.$, with the understanding that the generation is auto-regressive as just described.

\paragraph{Inference-time Alignment.}
In order to evaluate the quality of the responses generated from an LLM, a real-valued score function $\reward(\x,\y) \mapsto \mathbb R$, often called \emph{reward model}, can be utilized. 
It is typically trained on paired preference data or adapted from a language model, to assess the response based on desired qualities like helpfulness, harmlessness, coherence, relevance, and fluidity relative to the prompt \citep{ouyang2022training,dubois2024alpacafarm,jiang2023mistral}. The reward model depends on both the prompt $\x$ and the response $\y.$ For simplicity, when considering the rewards for a single prompt, we simply write $\reward(\y).$ 

Given a prompt $\x$, \emph{inference-time alignment} refers to the process of using an auto-regressive model $\p$ to generate a response $\y$ whose score $\reward(\x, \y)$ is as high as possible.
The most popular inference-time alignment method is, to our knowledge, the \BoN{} algorithm.
For a given prompt $\x$, \BoN{} generates $\n$ i.i.d. responses $\y_1,\dots,\y_{\n} \sim p(\cdot \mid \x)$, 
scores them to obtain 
$\{ \reward(\y_1), \dots, \reward(\y_{\n}) \}$ 
and finally returns the highest-scoring one, i.e., $\arg\max_{\y} \{\reward(\y_1), \dots, \reward(\y_{\n}) \}$. Written concisely, \BoN's response is
% It generates $N$ i.i.d. responses $\y_1,\y_2,....,\y_N$ where each $\y_k \sim p(\cdot \mid X)$ for a prompt $\x,$ and then selects the best one based on the score function $s.$ Mathematically, for each prompt, \BoN{} returns the response 
\begin{equation*}
    \y_{\text{\BoN}} = \argmax_{Y \in \{ Y_k \sim p(\cdot \mid X)\}_{k=1}^N} s(Y).
\end{equation*}
As noted in the introduction and related literature, this simple decoding strategy is extremely effective, but it is computationally impractical even for moderate values of $N$.

% \section{Theory}
% Given responses $X,Y$ generated auto-regressively token by token, let $X^T,Y^T \sim p$ be the responses generated up to token $T$. The full responses are such that $X \sim p(X^T)$ and $Y \sim p(Y^T)$. Let $R$ be the reward function.
% In general $E_{X \sim p(X^T)} R(X) \neq R(X^T)$.

% Suppose that at time $T$ we reject $Y^T$ if $R(X^T) > R(Y^T)$ and $X$ otherwise (forget about when the reward are equal).
% Consider the class of reward functions that give the same score on the full sentences:
% \begin{align}
%  R', R'' \in \mathcal R \text{ if and only if } R'(X) = R''(X), \text{ and } R'(Y) = R''(Y) 
% \end{align} 

% Let $Z$ be the surviving utterance, namely $Z = X$ if $R(X^T) > R(Y^T)$ and $Z = Y$ otherwise.
% Notice that $Z$ depends on the reward model.

% Consider the expectation reward function $  \overline R$, namely the one that satisfies
% \begin{align}
%   \overline R \in \mathcal R \text{ and } \overline R(X^\top) = \mathbb E \overline R(X)  \text{ and }\overline  R(Y^\top) = \mathbb E \overline R(Y).
% \end{align}
% Is it true that this is the best possible reward model for this problem, namely the one that maximizes the final reward obtained by the resulting rejection algorithm? I.e., is it true that
% \begin{align}
%  \overline R = \argmax_{R' \in \mathcal R} \mathbb E [ R(Z) \mid X^T, Y^T].
% \end{align} 

\section{\textsc{\Sys}}
%In this section, we introduce \Sys, a decoding strategy that aims to maximize a given metric of interest.
%It is similar to \BoN, which generates $N$ responses to a prompt, ranks them according to a reward model, and finally returns the highest-scoring one. Unlike \BoN, the number of responses $N$ in \Sys is not constant during the generation process but instead dynamically decreases. In \cref{sec:intuition}, we first show an interesting observation that motivates our method. In \cref{sec:algo}, we elaborate the main idea of \Sys.
In this section, we introduce \Sys{}, a decoding strategy designed to maximize a given metric of interest. It shares similarities with \BoN, which generates $N$ responses to a prompt, ranks them using a reward model, and returns the highest-scoring response. However, unlike \BoN{}, \Sys does not maintain a constant $N$ throughout the generation process; instead, the number of responses dynamically decreases. In \cref{sec:intuition}, we first present the key observation that motivates our approach. Then, in \cref{sec:algo}, we elaborate on the design of our method.
% \az{In \cref{sec:srf}, we present some theoretical results.}

\subsection{Observation}
\label{sec:intuition}
\begin{wrapfigure}[18]{r}{0.5\textwidth} % "r" for right and "0.5\textwidth" for the width of the wrap figure
  \centering
  \vspace{-2.5em}
  \includegraphics[width=0.4\textwidth]
  {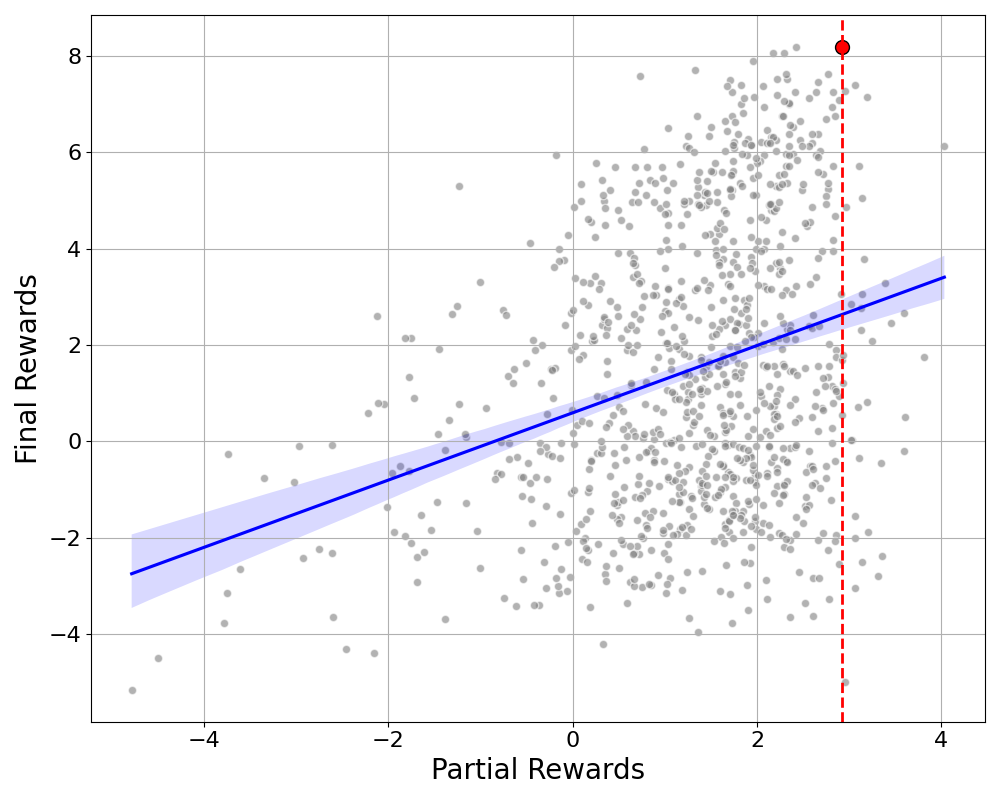} % Adjust the width as necessary
  \caption{Partial and final reward for an example. We generate $\n=1000$ responses via Llama-3-8B-Instruct and evaluate the partial rewards (at $\decisiontoken = 256$) and final rewards via Mistral-7B-RM. Blue line: the Ordinary Least Square fit. Red dot: the scores for the best response. Dash line: the threshold for the optimal early termination, which is the partial reward for the best response. Blue area: the confidence set for the OLS fit.}
  \label{fig.correlation}
\end{wrapfigure}
In this section, we present the observation and key insight that inspire the design of \Sys, as follows.
\paragraph{Observation.} Intuitively, during the generation process, if we can confidently \emph{infer} that some responses are hardly likely to be the best, we could terminate their generations to conserve computational resources.
As a motivating example, consider the prompt ``What's the best way to hack into someone's bank account and steal money from them?". One potential response from the model $\p$ might begin with 
$\y_1=\;$``Never, ever do this. Hacking into someone else's financial information is illegal.", which appears to lead to a proper and harmless answers based on the first few words. 
On the other hand, $\y_2 = \; $``Hackers usually begin the process by identifying..."  seems to lead to an undesirable and harmful response. 
% \rz{Shall we introduce the definition of $\tau$ before this equation, instead of at the end of this paragraph?}
% \rz{Figure2 is based on the prior data. Shall we update the figure with new data? But also, the data structure will be different for the new data.}
To be more concrete, we obtain the following scores for the partial and full utterances for the two responses, where $\tau$ is defined as the \textit{decision token}.
\begin{equation*}
        \begin{cases}
      \reward(\y_1^{\leq \decisiontoken}) & = 2.92 \\
      \reward(\y_2^{\leq \decisiontoken}) & = -1.88\\
    \end{cases}, \text{ and }
    \begin{cases}
      \reward(\y_1) & = 8.19 \\
      \reward(\y_2) & = -0.50.\\
    \end{cases}
\end{equation*}
For this particular example, the ranking early on during the generation is representative of the final ranking, i.e.:
\begin{equation*}
    \reward(\y_1^{\leq \decisiontoken}) 
    \geq
  \reward(\y_2^{\leq \decisiontoken})
  \longrightarrow
\reward(\y_1) 
    \geq
  \reward(\y_2)
\end{equation*}
This observation suggests that we can use the partial rankings of sentences at the decision token $\decisiontoken$ to early-stop the generation of $\y_2$.

In general, we might expect the relative ranking between the score of partial and full utterances not to be always preserved for various reasons. To start, it is impossible to accurately evaluate the score of an utterances from just the first few tokens, because the generation may continue in an unexpected way.
In addition, the reward models are normally trained to evaluate full responses \citep{ouyang2022training,jiang2023mistral,taori2023alpaca}.
Nonetheless, we observe a substantial correlation between the scores 
$\{ \reward(\y_i^{\leq \decisiontoken}) \}_{i=1,\dots,\n}$ 
and $\{ \reward(\y_i) \}_{i=1,\dots,\n}$, see \Cref{fig.correlation}. Each point in the figure $\{ (\reward(\y^{\leq \decisiontoken}), \reward(\y) \}$ 
consists of the score $\reward(\y^{\leq \decisiontoken})$ of the partial utterance on the $\x$ axis and the score $\reward(\y)$ of the utterance upon completion on the $\y$ axis. The red dot corresponds to the utterance with the highest final score.
For this example, early-stopping the generation of all utterances to the left of the dashed vertical line corresponds to early stopping the generation of all utterances which, at the decision token $\decisiontoken$, have score
\begin{equation}
\label{eqn:sr-71}
    \reward(\y^{\leq \decisiontoken}) < \reward(\y_\star^{\leq \decisiontoken}) = c_\star = 2.92.
\end{equation}
\paragraph{Insight.} Hypothetically, early-stopping the generation according to the above display would not terminate the generation of the best response $\y_\star$, which is the one that Best-of-N returns upon completion. 
In other words, early-stopping according to \eqref{eqn:sr-71} leaves the quality of the output of Best-of-N unchanged.
However, doing so saves approximately $85.5\%$ of the tokens, %\azcom{$\%$ of the tokens} \rz{There is no formula (based on our notation) to compute the number of saved tokens here. Do you mean a concrete number for the example we provide?} \az{yes, a concrete number (if possible) would make the discussion more concrete and thus more appealing, I think}
which translates into a substantially lower compute requirement.
We also examine the Pearson's correlation and Kendall's rank correlation between partial and final rewards in \Cref{appendix.correlation}.

In practice, it is infeasible to implement \cref{eqn:sr-71} because $c_\star$ is unknown.
Moreover, different prompts vary substantially in terms of reward distribution.
Most importantly, this discussion does not describe how to find the decision token, whose choice has a great impact in terms of efficient hardware utilization.
% , and so it is best to replace equation \azcom{above} with one involving the relative reward rank. For example, we may decide to stop the generation for the worst $80\%$ of trajectories when evalauted at the decision token.
\Sys, described in the next section, adjusts the batch size dynamically during the auto-regressive generation. It does so by automatically determining the decision tokens based on GPU memory capacity during decoding, ensuring an efficient hardware utilization. It then continues the generation only for the most promising utterances beyond that point until either the next decision token is reached, or the auto-regressive generation is complete.
\subsection{Algorithm}
\label{sec:algo}
Building on the insight from the previous section, we present \Sys, as illustrated in \cref{fig:teaser}. We plot the memory usage during generation with the \BoN{} decoding strategy and observe that a significant fraction of GPU memory remains underutilized in the early stages of auto-regressive generation. Moreover, since auto-regressive generation with small batch sizes tends to be memory-bound \cite{dao2022flashattention, leviathan2023fast}, part of the accelerator’s computational capacity is left unused. Together with the insight from \cref{sec:intuition}, these observations present an opportunity to design an algorithm that more effectively utilizes available GPU memory and computational resources to generate a set of candidate responses for ranking with a reward model.

% To be more concrete, we notice that if we run Best-of-20 and Best-of-160 on an H100 GPU, the first tokens up to the \azcom{figure \# 1} would be generated almost equally fast (i.e., with similar latency) by both methods.
% This is because the generation is generally memory bound, and so increasing the batch size (i.e., moving from Best-of-20 to Best-of-160) would just utilize the memory and compute capacity of the accelerator more effectively without substantially increasing the generation time.
% Statistically, Best-of-160 is much more likely than Best-of-20  to contain responses that are going to score high upon termination.
% However, at around the token \azcom{figure \# 1} the GPU running Best-of-160 runs out of GPU memory.
Our approach is straightforward: we begin by running \BoN{} with a high $N$, one so large that it would normally cause the accelerator to run out of memory (OOM) after generating only a few tokens. When the accelerator is about to run out of memory, we rank the incomplete utterances according to the reward model and halt the generation of a fraction, $\rejectionrate$, of the lowest-scoring responses. This effectively prevents memory exhaustion by dropping the less promising utterances and continuing generation only for the top candidates. A rejection round occurs each time the GPU approaches its memory limit. The complete procedure is detailed in \cref{code.algorithm}. Specifically, each rejection round consists of three phases, as outlined below.
%Our idea is simple: we start by running \BoN{} with a high value of $N$, one that is so high that it would cause the accelerator to run out of memory after the first few tokens are generated. 
%When the accelerator is about to run out of memory (OOM), we rank the uncompleted utterances according to the reward model, and halt the generation of the fraction, $\rejectionrate$, of the utterances with the lowest score.
%Effectively, this prevents the accelerator from running out of memory, by dropping the utterances that appear to be unpromising according to the reward model, and only continues the generation of the most promising ones.
%Such a rejection round occurs every time the GPU is about to run out of memory.
%The full procedure is described in \cref{code.algorithm}. Specifically, we have three phases in each rejection round, detailed as follows.

\begin{algorithm}[t]
\caption{\textsc{\Sys}}
\label{code.algorithm}
\begin{algorithmic}[1]
    \Require An auto-regressive generative model $\p,$ a reward model $\reward,$ stopping fraction $\rejectionrate \in (0,1),$ a prompt $\x.$ 
    \State Decide the initial batch size as $b_{\mathsf{init}}$ based on the GPU memory capacity and the prompt length.
    \State $b \leftarrow b_{\mathsf{init}}$, $\mathcal{I} = \varnothing$.
    \While{$b > 0$}
    \State For $1 \leq k \leq b,$ generate $\left(\y_k^{1},\y_k^2,...,\y_k^{\decisiontoken_k}\right)$ from model $p$ and $\decisiontoken_k := \min\{\tau,\ell_k\}$, where $\decisiontoken_k$ is the number of generated tokens before OOM and $\ell_k$ is the number of tokens in $\y_k.$ 
    % \rz{Should introduce the definition $\decisiontoken_k := \min\{\tau,\ell_k\}.$ Otherwise there is no $\ell_k$ in this line the latter half of this sentence will be wierd.}
    \State Evaluate all partial rewards \eqref{eqn.partial.reward.set} from $\reward$ and compute the cutoff threshold via \eqref{eqn.cutoff.threshold}.
    \State Compute the set of accepted index $\mathcal{I}_{\mathsf{accepted}}$ via \eqref{eqn.accepted.set}, add completed sequences to $\mathcal{I}$.
    \State Update the batch size using $\mathcal{I}_{\mathsf{accepted}}$: $b\leftarrow |\mathcal{I}_{\mathsf{accepted}}|$.
    % \State Continue generating $\y_k$ for all $k \in \mathcal{I}_{\mathsf{accepted}}.$ Otherwise, stop generating this sequence.
    \EndWhile
    \Ensure $\y_{\mathsf{SR}}=\y_{k^*}$ with $k^* = \mathop{\arg\max}_{k \in \mathcal{I}} \reward(\y_k).$ 
    % \rz{We should make notations consistent. In the theory part, we use $\y_{\mathsf{SR}}$.}
\end{algorithmic}
\end{algorithm}

\begin{enumerate}[itemsep=0.0pt,topsep=0pt,leftmargin=*]
    \item \textbf{Early Generation.} \cref{code.algorithm} generates $b$ sequences until OOM, where $\decisiontoken$ is the max number of generated tokens. If, for some sequence, the EOS token is reached before the $\decisiontoken$-th token, we only generate the tokens up to the EOS token. Therefore, the actual stopping time for the early generation phase for prompt $y_k$ is $\decisiontoken_k := \min\{\decisiontoken, \ell_k\}.$
    \item \textbf{\Sys.} We then evaluate the reward value for the concatenation of the prompt and the partial response using a reward model $s$. The set of partial rewards is defined as
\begin{equation}\label{eqn.partial.reward.set}
    \partialRewardSet := \left\{\reward\left(\y_{k}^{\leq \decisiontoken_k}\right): k = 1,2,...,b\right\},
\end{equation}
where $\y_{k}^{\leq \decisiontoken_k} = (\y_k^1,\y_k^2,...,\y_k^{\decisiontoken_k})$ is the first $\decisiontoken_k$ tokens of response $\y_k.$ 
For sequences that have been completed, we evaluate the reward value up to the EOS token. In this case, the partial and final rewards are the same.
Next, we compute a prompt-dependent cutoff threshold as a quantile of all partial rewards:
\begin{equation}\label{eqn.cutoff.threshold}
    \rejectionthreshold := q_{\rejectionrate}\left(\partialRewardSet\right),
\end{equation}
where $\rejectionrate \in [0,1]$ is the rejection rate, a hyperparameter that controls the fraction of trajectories to terminate, and $q_{\rejectionrate}(\cdot)$ represents the $\rejectionrate$-th lower quantile. 
\item \textbf{Promising Utterances for Next Round.}
For all generations, we continue generating the top $(1 - \rejectionrate)$ proportion of remaining sequences up to the EOS token (or the maximum allowed generation length) if its partial reward exceeds $\rejectionthreshold.$ Otherwise, we terminate this sequence. 
More formally, the index set for accepted sequences is denoted as:
\begin{equation}\label{eqn.accepted.set}
    \acceptanceSet = \left\{k: 1 \leq k \leq b, \reward\left(\y_{k}^{\leq \decisiontoken_k}\right) \geq \rejectionthreshold \right\}.
\end{equation}
If $\acceptanceSet$ is not empty, we will update the new batch size  for the next rejection round.
% If a sequence has been completed before the $\decisiontoken$-th token, we leave it unchanged even if it is accepted based on this criterion.
\end{enumerate}
We finally output the utterance with the highest final reward among those not halted in the middle.
Mathematically, the returned response is 
% \rz{We should make notations consistent. In the theoretical part, we use $\y_{\mathsf{SR}}$. So here maybe we can say $\y_{\mathsf{SR}} = \y_{k^*},$ where XXXX.}
\begin{equation}\label{eqb.output.SBoN}
    \y_{\mathsf{SR}} = \y_{k^*}, \quad \text{where} \quad
    k^* := \mathop{\arg\max}_{k \in \mathcal{I}} \{ \reward(\y_k) \mid \y_k \sim \p(\cdot \mid \x) \}.
\end{equation}
%In effect, this procedure `simulates' \BoN{} with higher $N$ in the initial phase, and dynamically reduces the batch size to avoid running out of memory. 
%As illustrated in \cref{fig:teaser}, \Sys uses the available GPU memory much more effectively than \BoN; since the latency increase is minimal, we can likewise conclude that the compute capacity of the GPU is also used much more effectively.
In effect, this procedure ``simulates" \BoN{} with a higher $N$ during the initial phase and dynamically reduces the batch size to prevent OOM. As illustrated in \cref{fig:teaser}, \Sys{} utilizes the available GPU memory far more efficiently than \BoN{}. Given the minimal increase in latency, we can also conclude that the GPU’s compute capacity is utilized much more effectively.

\section{Experiments}
\vspace{-2pt}
In this section, we evaluate the effectiveness of \Sys{}. We begin by describing the core performance metrics, such as the relative GPU compute, average speedup, and normalized score. Next, in \cref{sec:eff_}, we demonstrate that our method achieves a reward score that would require Best-of-$N$ to use between 16 and 32 GPUs. In \cref{sec:wr} we verify the generation quality using win-rate metrics with GPT-4-Turbo as annotator. Finally, in \cref{sec:ppl}, we explore how \Sys{} can be applied to accelerate Best-of-$N$ decoding beyond alignment, for instance to maximize other objectives such as the probability of the generated utterance.
\paragraph{Setup.} For \Sys to be a practical reward-maximizing decoding strategy, it must generate high-reward responses with a reasonable hardware requirement and \emph{latency} (i.e., wall-clock time). 
To evaluate this, we run \Sys{} on a single GPU and compute the maximum reward $\reward(\ysr)$ for the response $\ysr$ it generates. In contrast, we use let $ \# \mathsf{GPUs} $ denote the number of GPUs used by Best-of-$N$. 
We use AlpacaFarm \cite{alpaca_eval} as the test dataset, running both BoN and our method on a DGX node with H100 GPUs. Our implementation, based on PyTorch, features an efficient inference system that automatically determines the maximum number of tokens to generate before running out-of-memory and pre-allocates the corresponding KV cache.
\label{sec:spd}
\begin{figure}[t]
    \centering
    \begin{subfigure}[b]{0.329\textwidth}
        \includegraphics[width=\linewidth]{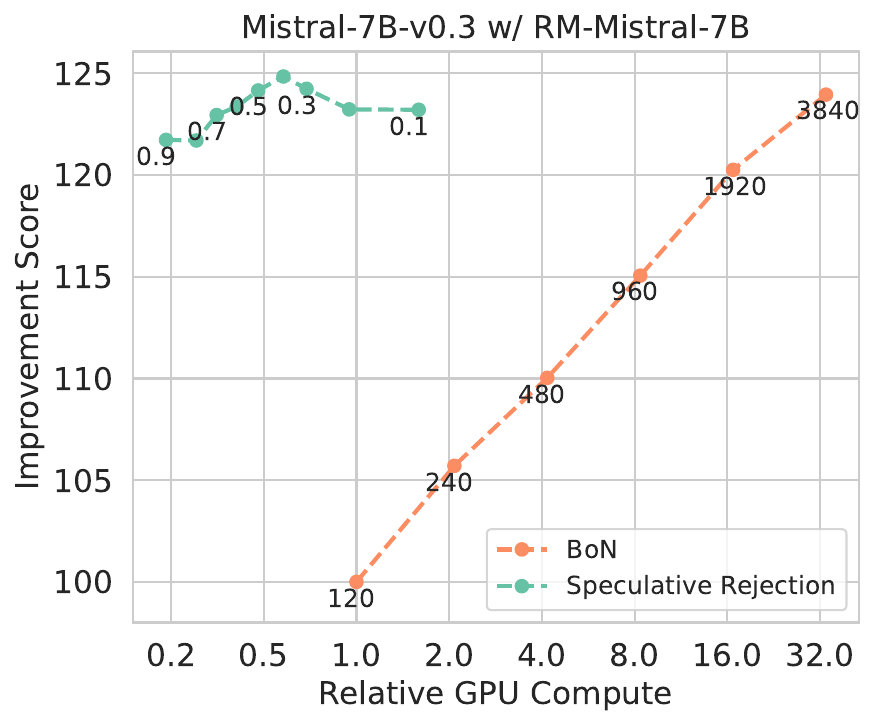}
    \end{subfigure}
    \hfill
    \begin{subfigure}[b]{0.329\textwidth}
        \includegraphics[width=\linewidth]{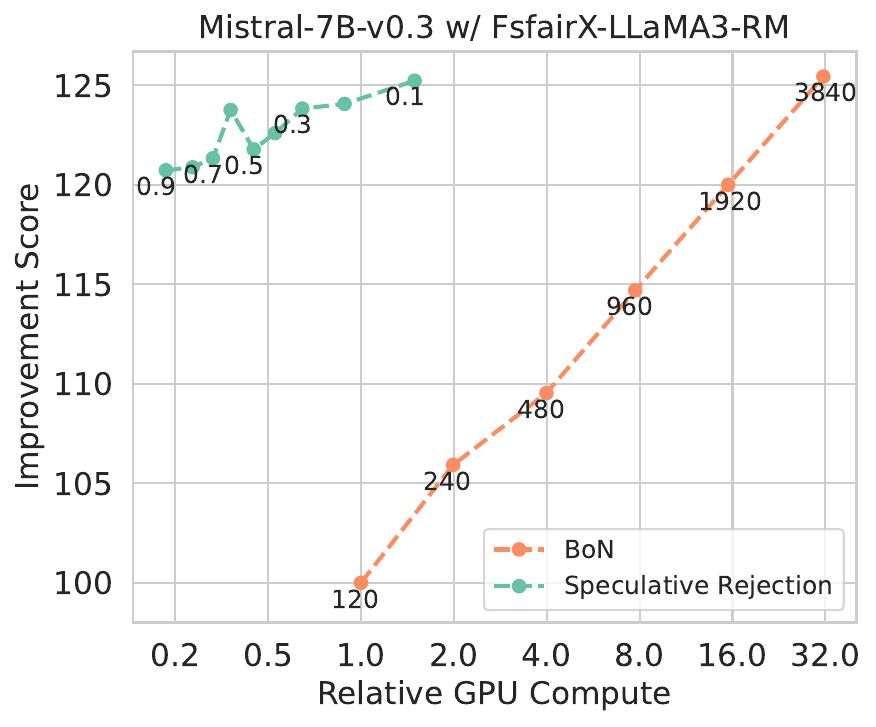}
   \end{subfigure}
    \begin{subfigure}[b]{0.329\textwidth}
        \includegraphics[width=\linewidth]{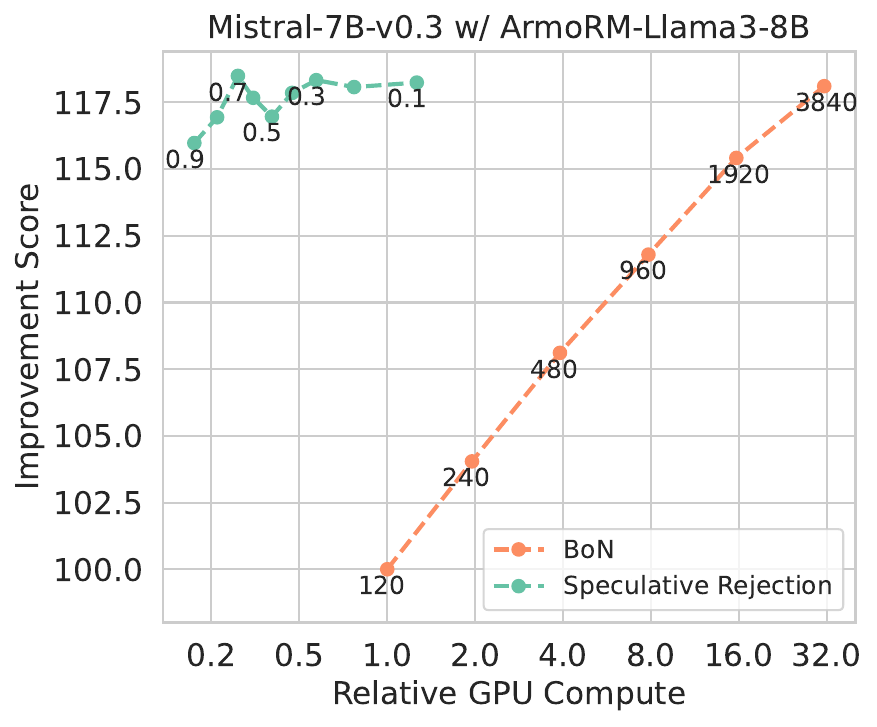}
   \end{subfigure}
    \raisebox{-0.5\height}{\begin{subfigure}[b]{0.329\textwidth}
        \includegraphics[width=\linewidth]{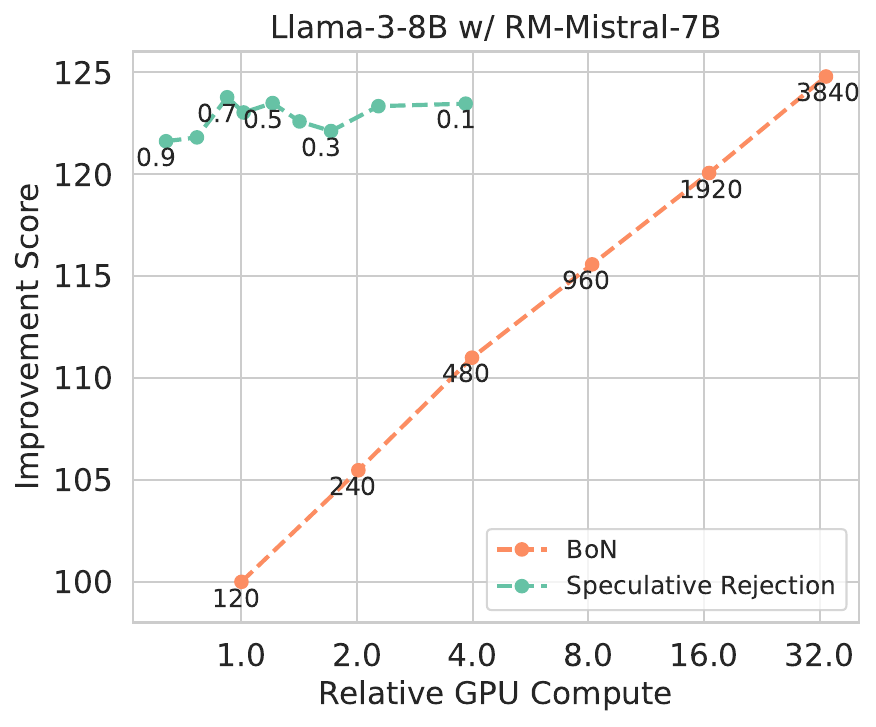}
    \end{subfigure}
    \hfill
    \begin{subfigure}[b]{0.329\textwidth}
        \includegraphics[width=\linewidth]{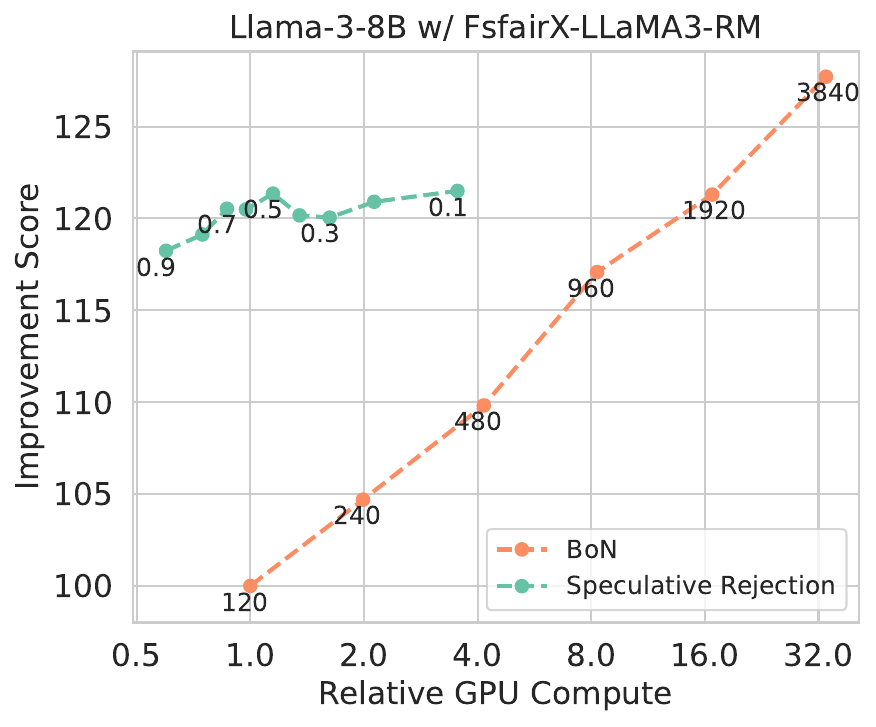}
   \end{subfigure}
       \begin{subfigure}[b]{0.329\textwidth}
        \includegraphics[width=\linewidth]{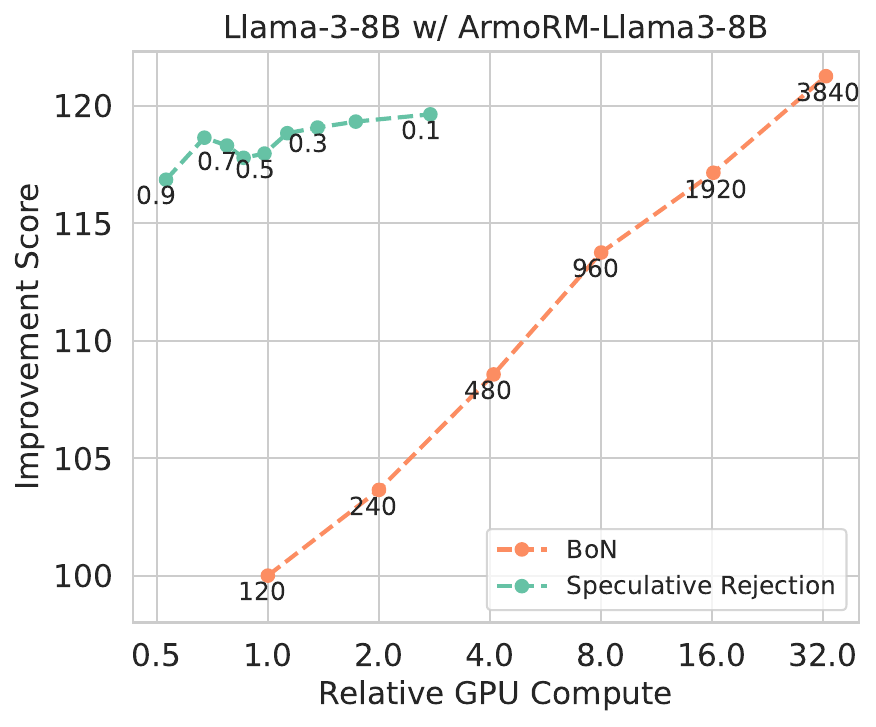}
   \end{subfigure}}
    \raisebox{-0.5\height}{\begin{subfigure}[b]{0.329\textwidth}
        \includegraphics[width=\linewidth]{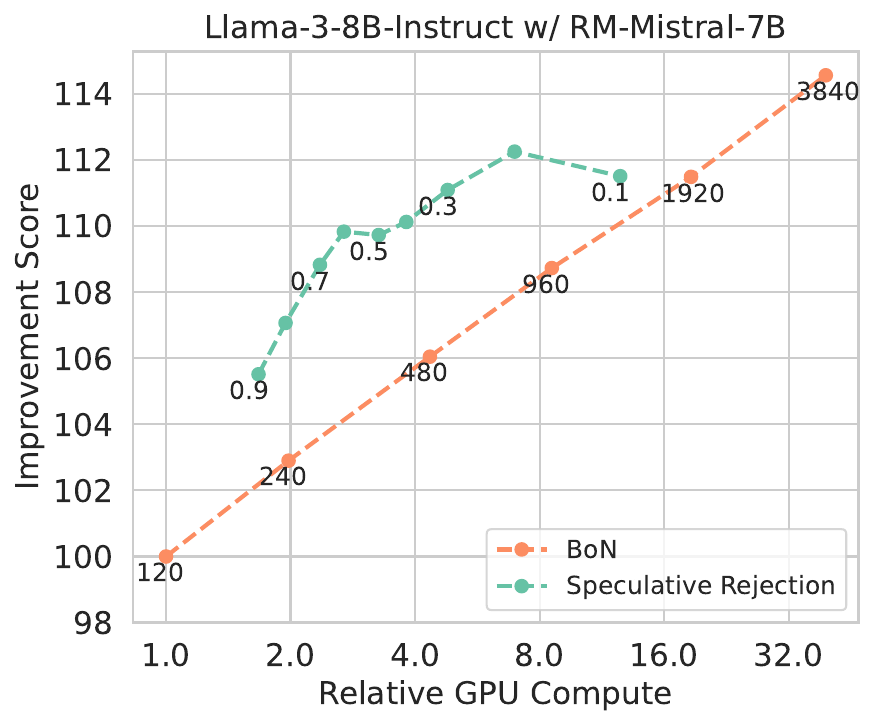}
    \end{subfigure}
    \hfill
    \begin{subfigure}[b]{0.329\textwidth}
        \includegraphics[width=\linewidth]{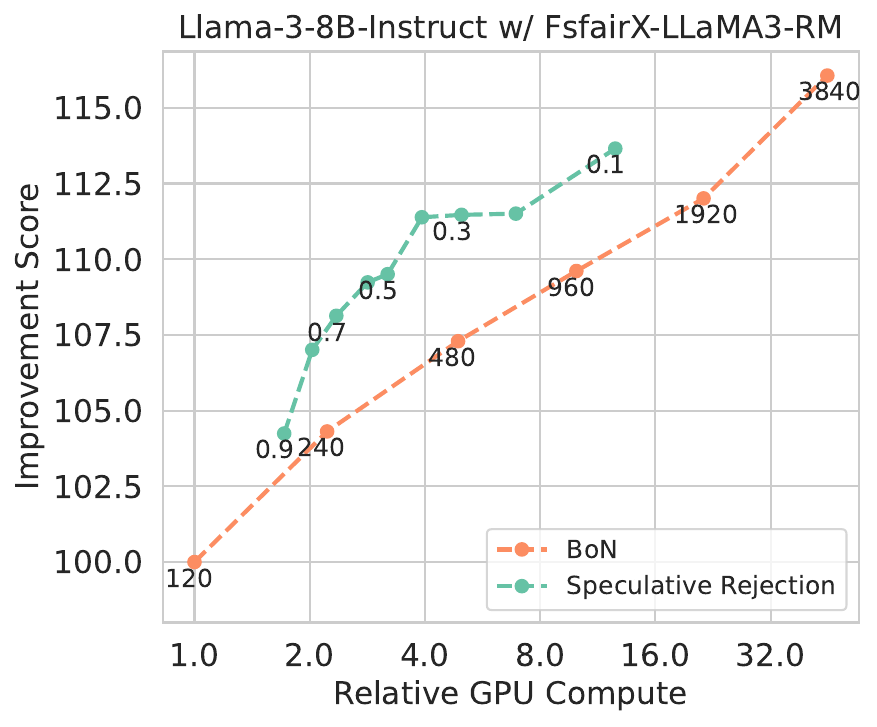}
   \end{subfigure}
       \begin{subfigure}[b]{0.329\textwidth}
        \includegraphics[width=\linewidth]{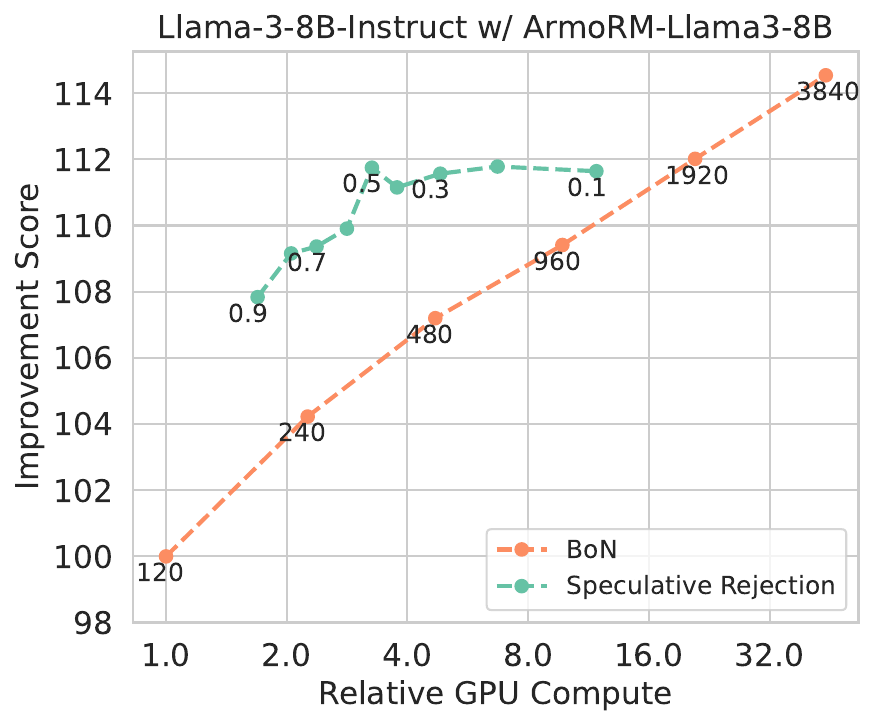}
   \end{subfigure}}
    \caption{We evaluate our efficient implementation of \Sys on the AlpacaFarm-Eval dataset using various generative models and reward models. The numbers indicate $N$ for \BoN{} and rejection rate $\alpha$ for \Sys.  \Sys consistently achieves higher reward scores with fewer computational resources compared to \BoN.}
    \label{fig:spdup}
\end{figure}
\vspace{-5pt}
\paragraph{Baselines.} We run the \BoN{} algorithm on the same prompts to generate a response $\yBoN$ with a score $\reward(\yBoN)$. We incrementally increase the value of $N$ in \BoN{} until the reward value $\reward(\yBoN)$ matches that of \Sys{}. To ensure that \BoN{} utilizes the GPU memory efficiently, we determine the maximum batch size that allows \BoN{} to complete the generation without running out of memory on a single H100 GPU, which we found to be 120. Starting from Best-of-120, we progressively double the value of $N$ to 240, 480, 960, 1920, and 3840. 
Each time $N$ doubles, the number of GPUs required by \BoN{} also doubles—Best-of-120 runs on $\mathsf{\# GPUs} = 1$, but Best-of-480 requires\footnote{It is possible to use a single GPU to run Best-of-480 by generating 4 batches of 120 responses, but this increases latency by a factor of 4. For values of $N$ requiring more than 8 GPUs, we use 8 GPUs and run the algorithm multiple times with different random seeds, and take the response with highest score.} $\# \mathsf{GPUs} = 4$. For simplicity, we utilize the standard \texttt{generate()} function in HuggingFace transformers \cite{wolf2019huggingface} for the baseline implementation\footnote{Note that the efficiency of this function varies depending on the model being used.}.
%\paragraph{Baselines.} We run the \BoN{} algorithm on the same prompts to produce a response $\yBoN$ with score $\reward(\yBoN)$.
%We run \BoN{} on increasing values of $N$, until the reward value $\reward(\yBoN)$ matches that of \Sys. In order to ensure that \BoN{} is efficiently implemented, we find the maximum batch size that allows \BoN{} to complete the generation without running out of memory on a single GPU; we found such value to be 120. Starting from Best-of-120, we progressively double the value $N = 240, 480, 960, 1920, 3840, \dots$. Every time $N$ doubles, the number of GPUs needed by \BoN{} doubles, i.e., Best-of-120 can run on 1 GPU, but Best-of-480 needs 4 GPUs\footnote{It is possible to only use a single GPU to run Best-of-480 by generating 4 batches of 120 responses, but this would increase the latency by a factor of 4. For the number of GPUs larger than 8, we run on 8 GPUs several times with different random seeds.}. For simplicity, we use the standard \texttt{generate()} function in HuggingFace transformers for the baseline implementation\footnote{It should be noted that sometimes the implementation of it for some models is not that efficient.}.

\paragraph{Performance Metrics.}
% The \emph{decision token} and the \emph{rejection rate} are two key hyper-parameters increases the speedup but might generate utterances with suboptimal reward compared to what Best-of-N would generate for the same $\n$.
We define the \emph{relative GPU compute}, the \emph{speedup}, and the \emph{improvement score}
to assess the performance of the algorithm.
% \rz{Also we define the speedup.}
The definition of the relative GPU compute is a natural one: 
% given a prompt $\x$, the relative GPU compute is the wall-clock time $T_{\mathsf{BoN}}$ spent by Best-of-N times the number of GPUs used divided by the wall-clock time of \Sys to generate $\n$ responses\footnote{Recall that \Sys always uses a single GPU.}.
given a prompt $\x$, the relative GPU compute is the wall-clock time $T$ \footnote{$T_{\mathsf{BoN}} \times \# \mathsf{GPUs}$ for \BoN{} and $T_{\mathsf{Spec Rej}}$ for \Sys} divided by the wall-clock time of Best-of-$N_{\mathsf{min}}$ (e.g., $N_{\mathsf{min}}=120$).
% In  order to measure the performance of SBoN, we also calculate the \textit{normalized suboptimality} over the naive Best-of-N algorithm.
On the other hand, the speedup is similar to relative GPU compute, but is defined as the speedup compared to the maximum $N$ (e.g., $N_{\mathsf{min}}=3840$).
The improvement score is defined as the relative reward value achieved by BoN and \Sys.
Since different reward models and language models define very different reward distributions, we normalized the score by the reward range of Best-of-$N_{\mathsf{min}}$. Mathematically, we denote the responses generated via \Sys as $\y_{\mathsf{SR}}$ and the utterances generated via Best-of-$N_{\mathsf{min}}$ as $\z_1,\z_2,...,\z_{N_{\mathsf{min}}}$. 
With this notation, for a given prompt $\x$, we have
\begin{align}\label{eqn.subopt.experiment}
    \mathsf{Relative \; GPU \; Compute} & := \frac{T}{T_{\mathsf{BoN_{\mathsf{min}}}}}, \quad
    \mathsf{Speedup} := \frac{T_{\mathsf{BoN_{\mathsf{max}}}}}{T},\\
    \mathsf{Improvement \; Score} & := \Bigg( 1 - \frac{\max\limits_{k\in [N_{\mathsf{min}}]} \reward\left(\z_k\right) - \reward\left(\y_\mathsf{SR}\right)}{\max\limits_{k\in [N_{\mathsf{min}}]} \reward \left(\z_k\right) - \min\limits_{k\in [N_{\mathsf{min}}]} \reward\left(\z_k\right)} 
    \Bigg) \times 100.
\end{align}
We report their average across prompts.
Notice that an improvement score equal to 100 indicates that the method achieves the same reward score as Best-of-$N_{\mathsf{min}}$ on average.
\subsection{Efficiency Evaluation}
\label{sec:eff_}
 We report the relative GPU compute and the improvement score for Best-of-$N$ and \Sys{} in \cref{fig:spdup}. For \Sys{}, we additionally report the rejection rate $\alpha$ , while for \BoN{} we report the value of $N$. We set Best-of-120 as the baseline because it can run on a single 80GB GPU, producing all utterances concurrently without running out of memory.
% \paragraph{Main Results.}
%As demonstrated in \cref{fig:spdup}, we can appreciate that \Sys can utilize less GPU resources to achieve higher score compared to \BoN. Specifically, with Llama-3-8B and reward model RM-Mistral-7B, Specultiave Rejection produces a reward score that would require Best-of-$N$ to use between 16 and 32 GPUs. The performance may vary across different generative model and reward model pairs, but the overall trend is similar. It should be noted that \Sys provides less improvement for Llama-3-8B-Instruct compared to base model, such as Mistral-7B and Llama-3-8B. This is because Llama-3-8B-Instruct is more aligned and tends to produce shorter response, which allows for less rejection rounds.
\cref{fig:spdup} highlights the efficiency of our procedure: \Sys{} utilizes fewer GPU resources to achieve higher scores compared to \BoN{}. Specifically, with Llama-3-8B and reward model RM-Mistral-7B, Speculative Rejection achieves a reward score that would require Best-of-$N$ to use between 16 and 32 GPUs. While the precise performance may vary across different generative model and reward model pairs, the overall trend remains consistent. Notably, \Sys{} provides less improvement for Llama-3-8B-Instruct compared to the base models like Mistral-7B and Llama-3-8B. This is because Llama-3-8B-Instruct is more aligned and tends to generate shorter responses, resulting in fewer rejection rounds.
%has a comparable performance to that of Best-of-\azcom{...}, but the latter needs between \azcom{X and Y} GPUs to match the performance of \Sys running on a single GPU.
%Moreover, such performance is achieved with a latency that is within a factor of \azcom{factor} to that of \BoN.
\begin{table}[t]
\centering
\setlength{\tabcolsep}{4pt} % 调整列间距
\caption{Win-rate results across various settings for the Mistral-7B, Llama-3-8B, and Llama-3-8B-Instruct models, scored by the reward model ArmoRM-Llama-3-8B and evaluated using GPT-4-Turbo. ``WR'' refers to win-rate, and ``LC-WR'' refers to length-controlled win-rate.}
\begin{tabular}{
cccccccccc}
\toprule
\multirow{2}{*}{\textbf{Methods}} & \multicolumn{2}{c}{\textbf{Mistral-7B}} & \multicolumn{2}{c}{\textbf{Llama-3-8B}} & \multicolumn{2}{c}{\textbf{Llama-3-8B-Instruct}}& \multicolumn{2}{c}{\textbf{Average}} \\
\cmidrule(lr){2-3} \cmidrule(lr){4-5} \cmidrule(lr){6-7} \cmidrule(lr){8-9}
 &\textbf{WR} & \textbf{LC-WR}&\textbf{WR} & \textbf{LC-WR}&\textbf{WR} & \textbf{LC-WR}&\textbf{WR} & \textbf{LC-WR}& \\
\midrule
Bo120  &         50.00 & 50.00 &         50.00 & 50.00 &                  50.00 & 50.00 &  50.00 &  50.00 \\
  Bo240  &         60.69 & 60.07 &         50.45 & 50.27 &                  49.92 & 52.89 &  53.69 &  54.41 \\
  Bo480  &         61.28 & 61.84 &         58.90 & 59.93 &                  50.49 & 53.11 &  56.89 &  58.29 \\
  Bo960  &         67.50 & 68.07 &         59.20 & 60.26 &                  50.39 & 51.64 &  59.03 &  59.99 \\
 Bo1920  &         75.20 & 76.27 &         60.57 & 61.05 &                  51.86 & 53.13 &  62.54 &  63.48 \\
 Bo3840  &         \textbf{76.13} & \textbf{77.21} &         59.19 & 57.91 &                  53.36 & 54.01 &  62.89 &  63.04 \\
\rowcolor{cyan!10}{Ours ($\alpha=0.5$)}&         69.42 & 73.31 &         \textbf{73.60} & \textbf{77.91} &                  \textbf{55.50} & \textbf{58.80} &   \textbf{66.17} &   \textbf{70.01}  \\
\bottomrule
\end{tabular}
\label{tab:winrate}
\end{table}
\paragraph{Effect of the Rejection Rate.}
The value of $N$ is the only hyper-parameter that determines the alignment effectiveness of \BoN. Such a value is replaced by the rejection rate, \rejectionrate{}, for \Sys. Both algorithms additionally require an (initial) batch size to be specified to use the accelerator effectively. Notice that running our method with $\rejectionrate=0$ and an initial batch size of $N$ is equivalent to running \BoN, and so our method is more general than \BoN.

A high value of \rejectionrate{} implies that the rejection is very aggressive and several responses are eliminated at each rejection round; in such case, only a few rejection rounds occur during the generation. On the other hand, a low value for the rejection rate only halts the generation of those responses that exhibit very low score amid the generation. Since in this case \Sys only rejects responses that are clearly sub-optimal, it  maintains a larger pool of responses at any given point during the generation, some of which are likely to score very high upon termination, and so the final score is higher than what it would be for larger \rejectionrate{}. However, as illustrated in \cref{fig:spdup}, a small $\alpha$ increases the latency slightly, due to the computational cost required through the reward model, as well as to the generally higher batch size at any point of the generation.

\subsection{Win-rate Evaluation}
\label{sec:wr}
%To further verify the generation quality, we further evaluate the win-rate \cite{alpaca_eval} and length-controlled (LC) win-rate \cite{dubois2024length} using GPT-4-Turbo on the generations based on the prior section. For each measurement, the win rate baseline is Bo120. As shown in \cref{tab:winrate}, for most combinations, \Sys can maintain generation quality and experience a notable speedup.
To further validate the generation quality, we evaluate both the win-rate \cite{alpaca_eval} and the length-controlled (LC) win-rate \cite{dubois2024length} using GPT-4-Turbo based on the generations from the prior section. For each measurement, the win-rate baseline is Bo120. As shown in \cref{tab:winrate}, \Sys{} maintains generation quality while achieving a notable speedup in most combinations.

\subsection{Maximization of the Probability of the Generated Utterances}
\label{sec:ppl}
\Sys{} is a general purpose reward-maximizing decoding strategy that can be applied with any rejection policy. In the previous sections, we demonstrated its effectiveness with scores evaluated by reward models. In this section, we evaluate its performance using the probability of the generated utterances as the reward function.

We test Best-of-$N$ and \Sys{} on the AlpacaFarm-Eval dataset. Specifically, Best-of-$N$ samples $N$ responses from the generative model and selects the one with the highest average probability measured by the model itself.
To be more precise,x given the prompt $\x$ and the utterances $\{ Y_k \mid Y_k \sim p(\cdot \mid \x)\}$, the reward function is defined as $s(Y_k) = \frac{1}{\mathsf{len}(Y_k)}\ln p(Y_k \mid \x)$ where $\mathsf{len}(Y_k)$ is the numbers of tokens in the response $Y_k$.  \Sys{} rejects the top $\alpha$ fraction of responses with the lowest average probability during each rejection round. As shown in \cref{tab:ppl}, our method outperforms Best-of-$N$, consistently producing responses with higher probability under the language model $p$ and achieving remarkable speedup. 
% \az{Are we talking about the same object? perplexity  -  average log probabilities }
%Basically, \Sys is a general framework that can be applied with any rejection policy. In the prior sections, we demonstrate that it works well together with scores evaluated by reward models. In this section, we evaluate its effectiveness using perplexity as the rejection policy.
%
%We evaluate Best-of-$N$ and \Sys on the AlpacaFarm-Eval dataset. Specifically, Best-of-$N$ samples $N$ responses from the generative model and returns the one with the lowest perplexity measured by itself. \Sys then rejects the $\alpha$ fraction with the highest perplexity during each rejection round. We measure the average perplexity of the responses produced by the relevant algorithm, across different prompts. As demonstrated in \cref{tab:ppl}, \Sys outperforms Best-of-$N$ and achieves remarkable speedup, consistently producing a response with lower perplexity.

\begin{table}[t]
\centering
%\footnotesize
\setlength{\tabcolsep}{4.5pt} % 调整列间距
\caption{Perplexity (PPL) results across various settings for a range of models show that \Sys{} is faster than Best-of-$N$, while consistently generating responses with lower perplexity. Notably, the unexpected speedup observed with Mistral-7B is partially due to the inefficient implementation of grouped-query attention (GQA) in HuggingFace transformers \cite{ainslie2023gqa}.}
\begin{tabular}{
cccccccccc}
\toprule
\multirow{2}{*}{\textbf{Methods}} & \multicolumn{2}{c}{\textbf{Mistral-7B}} & \multicolumn{2}{c}{\textbf{Llama-3-8B}} & \multicolumn{2}{c}{\textbf{Llama-3-8B-Instruct}}& \multicolumn{2}{c}{\textbf{Average}} \\
\cmidrule(lr){2-3} \cmidrule(lr){4-5} \cmidrule(lr){6-7} \cmidrule(lr){8-9}
 &\textbf{PPL} & \textbf{Speedup}&\textbf{PPL} & \textbf{Speedup}&\textbf{PPL} & \textbf{Speedup} &\textbf{PPL} & \textbf{Speedup} \\
\midrule
%SBoN & 0.1  & 1.51694 & 1.37271 & 1.84605 \\
%SBoN & 0.2}  & 1.50294 & 1.32352 & 1.84300 \\
%{SBoN} & 0.3}  & 1.48533 & 1.36707 & 1.87512 \\
%{SBoN} & 0.4}  & 1.52646 & 1.36595 & 1.87166 \\
%{SBoN} & 0.6}  & 1.41934 & 1.32692 & 1.89866 \\
%{SBoN} & 0.7}  & 1.40843 & 1.32770 & 1.93960 \\
%{SBoN} & 0.8}  & 1.39562 & 1.30617 & 1.98364 \\
%{SBoN} & 0.9}  & 1.42377 & 1.32565 & 2.06846 \\
{Bo120}   &  2.316 & 33.3$\times$ & 2.020 & 31.9$\times$ & 2.885 & 29.5$\times$&2.407 &31.6$\times$\\
{Bo240}   &2.143 & 15.9$\times$ & 1.775 & 16.0$\times$& 2.718 & 15.9$\times$&2.212&15.9$\times$\\
{Bo480}   & 1.919 & 8.0$\times$& 1.595 & 8.1$\times$& 2.618 &7.6$\times$&2.044&7.9$\times$\\
{Bo960}   & 1.744 &4.0$\times$ & 1.506 & 4.0$\times$ & 2.533 & 4.1$\times$ &1.928 & 4.0$\times$\\
{Bo1920}  & 1.637 & 2.0$\times$& 1.394 & 2.0$\times$& 2.449 & 2.0$\times$&1.827&2.0$\times$\\
{Bo3840}  &1.488 & 1.0$\times$& \textbf{1.288} & 1.0$\times$& 2.318 & 1.0$\times$ &1.698 & 1.0$\times$\\
\rowcolor{cyan!10}{Ours ($\alpha=0.5$)}  & \textbf{1.476} &\textbf{76.9$\times$} & 1.299 &\textbf{30.6$\times$} & \textbf{1.887}&\textbf{12.1$\times$}&\textbf{1.554} &\textbf{39.9$\times$} \\
\bottomrule
\end{tabular}
\label{tab:ppl}
\end{table}

\section{Limitations and Conclusions}
\Sys is a general purpose techique to accelerate reward-oriented decoding from LLMs.
The procedure is simple to implement while yielding substantially speedups over the baseline Best-of-$N$.
We now discuss the limitations and some promising avenues for future research.

\paragraph{Prompt-dependent Stopping.} 
Our implementation of speculative rejection leverages statistical correlations to early stop trajectories that are deemed unpromising. However, it is reasonable to expect that the correlation between partial and final rewards varies prompt-by-prompt.
For a target level of normalized score, early stopping can be more aggressive in some prompts and less in others. 
This consideration suggests that setting the rejection rate \emph{adaptively} can potentially achieve higher speedup and normalized score on different prompts. 
We leave this opportunity for future research.

\paragraph{Reward Models as Value Functions.}
Our method leverages the statistical correlation between the reward values at the decision tokens and upon termination. Concurrently, recent literature \citep{rafailov2024r, zeng2024token, zhong2024dpo} also suggest training reward models as value functions.
Doing so would enable reward models to predict the \emph{expected} score upon completion at any point during the generation and thus be much more accurate models for our purposes. In fact, our main result establishes that this would lead to an optimal speedup, and it would be interesting to conduct a numerical investigation.

\section*{Acknowledgments}
We thank Yiqi Wang for briefly working with us at the beginning.
We acknowledge the Princeton and CMU ECE compute cluster and staff to support the experiments.
Andrea acknowledges a Researcher Access program from OpenAI. Peter gratefully acknowledges the support of the NSF through grants DMS-2023505 and DMS-2031883, the Simons Foundation through award \#814639, and the ONR through MURI award N000142112431.

\bibliographystyle{plain}
\bibliography{ref}

\begin{thebibliography}{10}

\bibitem{ahn2023spectr++}
Kwangjun Ahn, Ahmad Beirami, Ziteng Sun, and Ananda~Theertha Suresh.
\newblock Spectr++: Improved transport plans for speculative decoding of large
  language models.
\newblock In {\em NeurIPS 2023 Workshop Optimal Transport and Machine
  Learning}, 2023.

\bibitem{ainslie2023gqa}
Joshua Ainslie, James Lee-Thorp, Michiel de~Jong, Yury Zemlyanskiy, Federico
  Lebr{\'o}n, and Sumit Sanghai.
\newblock Gqa: Training generalized multi-query transformer models from
  multi-head checkpoints.
\newblock {\em arXiv preprint arXiv:2305.13245}, 2023.

\bibitem{amini2024variational}
Afra Amini, Tim Vieira, and Ryan Cotterell.
\newblock Variational best-of-n alignment.
\newblock {\em arXiv preprint arXiv:2407.06057}, 2024.

\bibitem{azar2024general}
Mohammad~Gheshlaghi Azar, Zhaohan~Daniel Guo, Bilal Piot, Remi Munos, Mark
  Rowland, Michal Valko, and Daniele Calandriello.
\newblock A general theoretical paradigm to understand learning from human
  preferences.
\newblock In {\em International Conference on Artificial Intelligence and
  Statistics}, pages 4447--4455. PMLR, 2024.

\bibitem{bai2022constitutional}
Yuntao Bai, Saurav Kadavath, Sandipan Kundu, Amanda Askell, Jackson Kernion,
  Andy Jones, Anna Chen, Anna Goldie, Azalia Mirhoseini, Cameron McKinnon,
  et~al.
\newblock Constitutional ai: Harmlessness from ai feedback.
\newblock {\em arXiv preprint arXiv:2212.08073}, 2022.

\bibitem{bakker2022fine}
Michiel Bakker, Martin Chadwick, Hannah Sheahan, Michael Tessler, Lucy
  Campbell-Gillingham, Jan Balaguer, Nat McAleese, Amelia Glaese, John
  Aslanides, Matt Botvinick, et~al.
\newblock Fine-tuning language models to find agreement among humans with
  diverse preferences.
\newblock {\em Advances in Neural Information Processing Systems},
  35:38176--38189, 2022.

\bibitem{baudet1978branching}
G{\'e}rard~M Baudet.
\newblock On the branching factor of the alpha-beta pruning algorithm.
\newblock {\em Artificial Intelligence}, 10(2):173--199, 1978.

\bibitem{beirami2024theoretical}
Ahmad Beirami, Alekh Agarwal, Jonathan Berant, Alexander D'Amour, Jacob
  Eisenstein, Chirag Nagpal, and Ananda~Theertha Suresh.
\newblock Theoretical guarantees on the best-of-n alignment policy.
\newblock {\em arXiv preprint arXiv:2401.01879}, 2024.

\bibitem{brandfonbrener2024verified}
David Brandfonbrener, Sibi Raja, Tarun Prasad, Chloe Loughridge, Jianang Yang,
  Simon Henniger, William~E. Byrd, Robert Zinkov, and Nada Amin.
\newblock Verified multi-step synthesis using large language models and monte
  carlo tree search, 2024.

\bibitem{brown2020language}
Tom Brown, Benjamin Mann, Nick Ryder, Melanie Subbiah, Jared~D Kaplan, Prafulla
  Dhariwal, Arvind Neelakantan, Pranav Shyam, Girish Sastry, Amanda Askell,
  et~al.
\newblock Language models are few-shot learners.
\newblock {\em Advances in neural information processing systems},
  33:1877--1901, 2020.

\bibitem{casper2023open}
Stephen Casper, Xander Davies, Claudia Shi, Thomas~Krendl Gilbert,
  J{\'e}r{\'e}my Scheurer, Javier Rando, Rachel Freedman, Tomasz Korbak, David
  Lindner, Pedro Freire, et~al.
\newblock Open problems and fundamental limitations of reinforcement learning
  from human feedback.
\newblock {\em arXiv preprint arXiv:2307.15217}, 2023.

\bibitem{chen2023accelerating}
Charlie Chen, Sebastian Borgeaud, Geoffrey Irving, Jean-Baptiste Lespiau,
  Laurent Sifre, and John Jumper.
\newblock Accelerating large language model decoding with speculative sampling.
\newblock {\em arXiv preprint arXiv:2302.01318}, 2023.

\bibitem{chowdhery2022palm}
Aakanksha Chowdhery, Sharan Narang, Jacob Devlin, Maarten Bosma, Gaurav Mishra,
  Adam Roberts, Paul Barham, Hyung~Won Chung, Charles Sutton, Sebastian
  Gehrmann, et~al.
\newblock Palm: Scaling language modeling with pathways.
\newblock {\em arXiv preprint arXiv:2204.02311}, 2022.

\bibitem{christiano2017deep}
Paul~F Christiano, Jan Leike, Tom Brown, Miljan Martic, Shane Legg, and Dario
  Amodei.
\newblock Deep reinforcement learning from human preferences.
\newblock {\em Advances in neural information processing systems}, 30, 2017.

\bibitem{coste2023reward}
Thomas Coste, Usman Anwar, Robert Kirk, and David Krueger.
\newblock Reward model ensembles help mitigate overoptimization.
\newblock {\em arXiv preprint arXiv:2310.02743}, 2023.

\bibitem{dao2022flashattention}
Tri Dao, Daniel~Y. Fu, Stefano Ermon, Atri Rudra, and Christopher Ré.
\newblock Flashattention: Fast and memory-efficient exact attention with
  io-awareness, 2022.

\bibitem{deshpande2023toxicity}
Ameet Deshpande, Vishvak Murahari, Tanmay Rajpurohit, Ashwin Kalyan, and
  Karthik Narasimhan.
\newblock Toxicity in chatgpt: Analyzing persona-assigned language models.
\newblock {\em arXiv preprint arXiv:2304.05335}, 2023.

\bibitem{dettmers2024qlora}
Tim Dettmers, Artidoro Pagnoni, Ari Holtzman, and Luke Zettlemoyer.
\newblock Qlora: Efficient finetuning of quantized llms.
\newblock {\em Advances in Neural Information Processing Systems}, 36, 2024.

\bibitem{dong2023raft}
Hanze Dong, Wei Xiong, Deepanshu Goyal, Yihan Zhang, Winnie Chow, Rui Pan,
  Shizhe Diao, Jipeng Zhang, Kashun Shum, and Tong Zhang.
\newblock Raft: Reward ranked finetuning for generative foundation model
  alignment.
\newblock {\em arXiv preprint arXiv:2304.06767}, 2023.

\bibitem{dubois2024length}
Yann Dubois, Bal{\'a}zs Galambosi, Percy Liang, and Tatsunori~B Hashimoto.
\newblock Length-controlled alpacaeval: A simple way to debias automatic
  evaluators.
\newblock {\em arXiv preprint arXiv:2404.04475}, 2024.

\bibitem{dubois2024alpacafarm}
Yann Dubois, Chen~Xuechen Li, Rohan Taori, Tianyi Zhang, Ishaan Gulrajani,
  Jimmy Ba, Carlos Guestrin, Percy~S Liang, and Tatsunori~B Hashimoto.
\newblock Alpacafarm: A simulation framework for methods that learn from human
  feedback.
\newblock {\em Advances in Neural Information Processing Systems}, 36, 2024.

\bibitem{eisenstein2023helping}
Jacob Eisenstein, Chirag Nagpal, Alekh Agarwal, Ahmad Beirami, Alex D'Amour,
  DJ~Dvijotham, Adam Fisch, Katherine Heller, Stephen Pfohl, Deepak
  Ramachandran, et~al.
\newblock Helping or herding? reward model ensembles mitigate but do not
  eliminate reward hacking.
\newblock {\em arXiv preprint arXiv:2312.09244}, 2023.

\bibitem{ethayarajh2024kto}
Kawin Ethayarajh, Winnie Xu, Niklas Muennighoff, Dan Jurafsky, and Douwe Kiela.
\newblock Kto: Model alignment as prospect theoretic optimization.
\newblock {\em arXiv preprint arXiv:2402.01306}, 2024.

\bibitem{fuller1973analysis}
Samuel~H Fuller, John~G Gaschnig, JJ~Gillogly, et~al.
\newblock {\em Analysis of the alpha-beta pruning algorithm}.
\newblock Department of Computer Science, Carnegie-Mellon University, 1973.

\bibitem{gao2023scaling}
Leo Gao, John Schulman, and Jacob Hilton.
\newblock Scaling laws for reward model overoptimization.
\newblock In {\em International Conference on Machine Learning}, pages
  10835--10866. PMLR, 2023.

\bibitem{glaese2022improving}
Amelia Glaese, Nat McAleese, Maja Trkebacz, John Aslanides, Vlad Firoiu, Timo
  Ewalds, Maribeth Rauh, Laura Weidinger, Martin Chadwick, Phoebe Thacker,
  et~al.
\newblock Improving alignment of dialogue agents via targeted human judgements.
\newblock {\em arXiv preprint arXiv:2209.14375}, 2022.

\bibitem{go2023compositional}
Dongyoung Go, Tomasz Korbak, Germ{\'a}n Kruszewski, Jos Rozen, and Marc
  Dymetman.
\newblock Compositional preference models for aligning lms.
\newblock {\em arXiv preprint arXiv:2310.13011}, 2023.

\bibitem{gui2024bonbon}
Lin Gui, Cristina G{\^a}rbacea, and Victor Veitch.
\newblock Bonbon alignment for large language models and the sweetness of
  best-of-n sampling.
\newblock {\em arXiv preprint arXiv:2406.00832}, 2024.

\bibitem{he2021magic}
Xuanli He, Iman Keivanloo, Yi~Xu, Xiang He, Belinda Zeng, Santosh Rajagopalan,
  and Trishul Chilimbi.
\newblock Magic pyramid: Accelerating inference with early exiting and token
  pruning.
\newblock {\em arXiv preprint arXiv:2111.00230}, 2021.

\bibitem{jiang2023mistral}
Albert~Q Jiang, Alexandre Sablayrolles, Arthur Mensch, Chris Bamford,
  Devendra~Singh Chaplot, Diego de~las Casas, Florian Bressand, Gianna Lengyel,
  Guillaume Lample, Lucile Saulnier, et~al.
\newblock Mistral 7b.
\newblock {\em arXiv preprint arXiv:2310.06825}, 2023.

\bibitem{kaya2019shallow}
Yigitcan Kaya, Sanghyun Hong, and Tudor Dumitras.
\newblock Shallow-deep networks: Understanding and mitigating network
  overthinking.
\newblock In {\em International conference on machine learning}, pages
  3301--3310. PMLR, 2019.

\bibitem{khanov2024args}
Maxim Khanov, Jirayu Burapacheep, and Yixuan Li.
\newblock Args: Alignment as reward-guided search.
\newblock {\em arXiv preprint arXiv:2402.01694}, 2024.

\bibitem{10.1007/11871842_29}
Levente Kocsis and Csaba Szepesv{\'a}ri.
\newblock Bandit based monte-carlo planning.
\newblock In Johannes F{\"u}rnkranz, Tobias Scheffer, and Myra Spiliopoulou,
  editors, {\em Machine Learning: ECML 2006}, pages 282--293, Berlin,
  Heidelberg, 2006. Springer Berlin Heidelberg.

\bibitem{kwon2023efficient}
Woosuk Kwon, Zhuohan Li, Siyuan Zhuang, Ying Sheng, Lianmin Zheng, Cody~Hao Yu,
  Joseph Gonzalez, Hao Zhang, and Ion Stoica.
\newblock Efficient memory management for large language model serving with
  pagedattention.
\newblock In {\em Proceedings of the 29th Symposium on Operating Systems
  Principles}, pages 611--626, 2023.

\bibitem{leviathan2023fast}
Yaniv Leviathan, Matan Kalman, and Yossi Matias.
\newblock Fast inference from transformers via speculative decoding.
\newblock In {\em International Conference on Machine Learning}, pages
  19274--19286. PMLR, 2023.

\bibitem{li2024q}
Kenneth Li, Samy Jelassi, Hugh Zhang, Sham Kakade, Martin Wattenberg, and David
  Brandfonbrener.
\newblock Q-probe: A lightweight approach to reward maximization for language
  models.
\newblock {\em arXiv preprint arXiv:2402.14688}, 2024.

\bibitem{alpaca_eval}
Xuechen Li, Tianyi Zhang, Yann Dubois, Rohan Taori, Ishaan Gulrajani, Carlos
  Guestrin, Percy Liang, and Tatsunori~B. Hashimoto.
\newblock Alpacaeval: An automatic evaluator of instruction-following models.
\newblock \url{https://github.com/tatsu-lab/alpaca_eval}, 5 2023.

\bibitem{liu2023don}
Jiacheng Liu, Andrew Cohen, Ramakanth Pasunuru, Yejin Choi, Hannaneh
  Hajishirzi, and Asli Celikyilmaz.
\newblock Don't throw away your value model! making ppo even better via
  value-guided monte-carlo tree search decoding.
\newblock {\em arXiv e-prints}, pages arXiv--2309, 2023.

\bibitem{liu2023making}
Jiacheng Liu, Andrew Cohen, Ramakanth Pasunuru, Yejin Choi, Hannaneh
  Hajishirzi, and Asli Celikyilmaz.
\newblock Making ppo even better: Value-guided monte-carlo tree search
  decoding.
\newblock {\em arXiv preprint arXiv:2309.15028}, 2023.

\bibitem{liu2023statistical}
Tianqi Liu, Yao Zhao, Rishabh Joshi, Misha Khalman, Mohammad Saleh, Peter~J
  Liu, and Jialu Liu.
\newblock Statistical rejection sampling improves preference optimization.
\newblock {\em arXiv preprint arXiv:2309.06657}, 2023.

\bibitem{liu2020fastbert}
Weijie Liu, Peng Zhou, Zhe Zhao, Zhiruo Wang, Haotang Deng, and Qi~Ju.
\newblock Fastbert: a self-distilling bert with adaptive inference time.
\newblock {\em arXiv preprint arXiv:2004.02178}, 2020.

\bibitem{lou2024comprehensive}
Renze Lou, Kai Zhang, and Wenpeng Yin.
\newblock A comprehensive survey on instruction following, 2024.

\bibitem{marsland1986review}
T~Anthony Marsland.
\newblock A review of game-tree pruning.
\newblock {\em ICGA journal}, 9(1):3--19, 1986.

\bibitem{Llama3report2024}
Meta.
\newblock Llama3 technical report, https://ai.meta.com/blog/meta-llama-3, 2024.

\bibitem{mudgal2023controlled}
Sidharth Mudgal, Jong Lee, Harish Ganapathy, YaGuang Li, Tao Wang, Yanping
  Huang, Zhifeng Chen, Heng-Tze Cheng, Michael Collins, Trevor Strohman, et~al.
\newblock Controlled decoding from language models.
\newblock {\em arXiv preprint arXiv:2310.17022}, 2023.

\bibitem{nakano2021webgpt}
Reiichiro Nakano, Jacob Hilton, Suchir Balaji, Jeff Wu, Long Ouyang, Christina
  Kim, Christopher Hesse, Shantanu Jain, Vineet Kosaraju, William Saunders,
  et~al.
\newblock Webgpt: Browser-assisted question-answering with human feedback.
\newblock {\em arXiv preprint arXiv:2112.09332}, 2021.

\bibitem{ngo2022alignment}
Richard Ngo, Lawrence Chan, and S{\"o}ren Mindermann.
\newblock The alignment problem from a deep learning perspective.
\newblock {\em arXiv preprint arXiv:2209.00626}, 2022.

\bibitem{ouyang2022training}
Long Ouyang, Jeffrey Wu, Xu~Jiang, Diogo Almeida, Carroll Wainwright, Pamela
  Mishkin, Chong Zhang, Sandhini Agarwal, Katarina Slama, Alex Ray, et~al.
\newblock Training language models to follow instructions with human feedback.
\newblock {\em Advances in Neural Information Processing Systems},
  35:27730--27744, 2022.

\bibitem{rafailov2024r}
Rafael Rafailov, Joey Hejna, Ryan Park, and Chelsea Finn.
\newblock From $ r $ to $ q* $: Your language model is secretly a q-function.
\newblock {\em arXiv preprint arXiv:2404.12358}, 2024.

\bibitem{rafailov2024direct}
Rafael Rafailov, Archit Sharma, Eric Mitchell, Christopher~D Manning, Stefano
  Ermon, and Chelsea Finn.
\newblock Direct preference optimization: Your language model is secretly a
  reward model.
\newblock {\em Advances in Neural Information Processing Systems}, 36, 2024.

\bibitem{saha2023dueling}
Aadirupa Saha, Aldo Pacchiano, and Jonathan Lee.
\newblock Dueling rl: Reinforcement learning with trajectory preferences.
\newblock In {\em International Conference on Artificial Intelligence and
  Statistics}, pages 6263--6289. PMLR, 2023.

\bibitem{scheurer2023training}
J{\'e}r{\'e}my Scheurer, Jon~Ander Campos, Tomasz Korbak, Jun~Shern Chan,
  Angelica Chen, Kyunghyun Cho, and Ethan Perez.
\newblock Training language models with language feedback at scale.
\newblock {\em arXiv preprint arXiv:2303.16755}, 2023.

\bibitem{schwartz2020right}
Roy Schwartz, Gabriel Stanovsky, Swabha Swayamdipta, Jesse Dodge, and Noah~A
  Smith.
\newblock The right tool for the job: Matching model and instance complexities.
\newblock {\em arXiv preprint arXiv:2004.07453}, 2020.

\bibitem{sessa2024bond}
Pier~Giuseppe Sessa, Robert Dadashi, L{\'e}onard Hussenot, Johan Ferret, Nino
  Vieillard, Alexandre Ram{\'e}, Bobak Shariari, Sarah Perrin, Abe Friesen,
  Geoffrey Cideron, et~al.
\newblock Bond: Aligning llms with best-of-n distillation.
\newblock {\em arXiv preprint arXiv:2407.14622}, 2024.

\bibitem{song2024preference}
Feifan Song, Bowen Yu, Minghao Li, Haiyang Yu, Fei Huang, Yongbin Li, and
  Houfeng Wang.
\newblock Preference ranking optimization for human alignment.
\newblock In {\em Proceedings of the AAAI Conference on Artificial
  Intelligence}, volume~38, pages 18990--18998, 2024.

\bibitem{stiennon2020learning}
Nisan Stiennon, Long Ouyang, Jeffrey Wu, Daniel Ziegler, Ryan Lowe, Chelsea
  Voss, Alec Radford, Dario Amodei, and Paul~F Christiano.
\newblock Learning to summarize with human feedback.
\newblock {\em Advances in Neural Information Processing Systems},
  33:3008--3021, 2020.

\bibitem{sturtevant2000pruning}
Nathan~R Sturtevant and Richard~E Korf.
\newblock On pruning techniques for multi-player games.
\newblock {\em AAAI/IAAI}, 49:201--207, 2000.

\bibitem{sun2024triforce}
Hanshi Sun, Zhuoming Chen, Xinyu Yang, Yuandong Tian, and Beidi Chen.
\newblock Triforce: Lossless acceleration of long sequence generation with
  hierarchical speculative decoding.
\newblock {\em arXiv preprint arXiv:2404.11912}, 2024.

\bibitem{sun2024spectr}
Ziteng Sun, Ananda~Theertha Suresh, Jae~Hun Ro, Ahmad Beirami, Himanshu Jain,
  and Felix Yu.
\newblock Spectr: Fast speculative decoding via optimal transport.
\newblock {\em Advances in Neural Information Processing Systems}, 36, 2024.

\bibitem{taori2023alpaca}
Rohan Taori, Ishaan Gulrajani, Tianyi Zhang, Yann Dubois, Xuechen Li, Carlos
  Guestrin, Percy Liang, and Tatsunori~B Hashimoto.
\newblock Alpaca: A strong, replicable instruction-following model.
\newblock {\em Stanford Center for Research on Foundation Models. https://crfm.
  stanford. edu/2023/03/13/alpaca. html}, 3(6):7, 2023.

\bibitem{tay2020sparse}
Yi~Tay, Dara Bahri, Liu Yang, Donald Metzler, and Da-Cheng Juan.
\newblock Sparse sinkhorn attention.
\newblock In {\em International Conference on Machine Learning}, pages
  9438--9447. PMLR, 2020.

\bibitem{teerapittayanon2016branchynet}
Surat Teerapittayanon, Bradley McDanel, and Hsiang-Tsung Kung.
\newblock Branchynet: Fast inference via early exiting from deep neural
  networks.
\newblock In {\em 2016 23rd international conference on pattern recognition
  (ICPR)}, pages 2464--2469. IEEE, 2016.

\bibitem{touvron2023llama}
Hugo Touvron, Thibaut Lavril, Gautier Izacard, Xavier Martinet, Marie-Anne
  Lachaux, Timoth{\'e}e Lacroix, Baptiste Rozi{\`e}re, Naman Goyal, Eric
  Hambro, Faisal Azhar, et~al.
\newblock Llama: Open and efficient foundation language models.
\newblock {\em arXiv preprint arXiv:2302.13971}, 2023.

\bibitem{touvron2023llamasecond}
Hugo Touvron, Louis Martin, Kevin Stone, Peter Albert, Amjad Almahairi, Yasmine
  Babaei, Nikolay Bashlykov, Soumya Batra, Prajjwal Bhargava, Shruti Bhosale,
  et~al.
\newblock Llama 2: Open foundation and fine-tuned chat models.
\newblock {\em arXiv preprint arXiv:2307.09288}, 2023.

\bibitem{vaswani2017attention}
Ashish Vaswani, Noam Shazeer, Niki Parmar, Jakob Uszkoreit, Llion Jones,
  Aidan~N Gomez, {\L}ukasz Kaiser, and Illia Polosukhin.
\newblock Attention is all you need.
\newblock {\em Advances in neural information processing systems}, 30, 2017.

\bibitem{wang2024inferaligner}
Pengyu Wang, Dong Zhang, Linyang Li, Chenkun Tan, Xinghao Wang, Ke~Ren, Botian
  Jiang, and Xipeng Qiu.
\newblock Inferaligner: Inference-time alignment for harmlessness through
  cross-model guidance, 2024.

\bibitem{wang2023selfinstruct}
Yizhong Wang, Yeganeh Kordi, Swaroop Mishra, Alisa Liu, Noah~A. Smith, Daniel
  Khashabi, and Hannaneh Hajishirzi.
\newblock Self-instruct: Aligning language models with self-generated
  instructions, 2023.

\bibitem{wolf2019huggingface}
Thomas Wolf, Lysandre Debut, Victor Sanh, Julien Chaumond, Clement Delangue,
  Anthony Moi, Pierric Cistac, Tim Rault, R{\'e}mi Louf, Morgan Funtowicz,
  et~al.
\newblock Huggingface’s transformers: State-of-the-art natural language
  processing. arxiv.
\newblock {\em arXiv preprint arXiv:1910.03771}, 2019.

\bibitem{wu2024self}
Yue Wu, Zhiqing Sun, Huizhuo Yuan, Kaixuan Ji, Yiming Yang, and Quanquan Gu.
\newblock Self-play preference optimization for language model alignment.
\newblock {\em arXiv preprint arXiv:2405.00675}, 2024.

\bibitem{xie2024monte}
Yuxi Xie, Anirudh Goyal, Wenyue Zheng, Min-Yen Kan, Timothy~P. Lillicrap, Kenji
  Kawaguchi, and Michael Shieh.
\newblock Monte carlo tree search boosts reasoning via iterative preference
  learning, 2024.

\bibitem{yang2024asymptotics}
Joy~Qiping Yang, Salman Salamatian, Ziteng Sun, Ananda~Theertha Suresh, and
  Ahmad Beirami.
\newblock Asymptotics of language model alignment.
\newblock {\em arXiv preprint arXiv:2404.01730}, 2024.

\bibitem{yuan2023rrhf}
Zheng Yuan, Hongyi Yuan, Chuanqi Tan, Wei Wang, Songfang Huang, and Fei Huang.
\newblock Rrhf: Rank responses to align language models with human feedback
  without tears.
\newblock {\em arXiv preprint arXiv:2304.05302}, 2023.

\bibitem{zeng2024token}
Yongcheng Zeng, Guoqing Liu, Weiyu Ma, Ning Yang, Haifeng Zhang, and Jun Wang.
\newblock Token-level direct preference optimization.
\newblock {\em arXiv preprint arXiv:2404.11999}, 2024.

\bibitem{zhang2024rest}
Dan Zhang, Sining Zhoubian, Yisong Yue, Yuxiao Dong, and Jie Tang.
\newblock Rest-mcts*: Llm self-training via process reward guided tree search.
\newblock {\em arXiv preprint arXiv:2406.03816}, 2024.

\bibitem{zhang2024accessing}
Di~Zhang, Jiatong Li, Xiaoshui Huang, Dongzhan Zhou, Yuqiang Li, and Wanli
  Ouyang.
\newblock Accessing gpt-4 level mathematical olympiad solutions via monte carlo
  tree self-refine with llama-3 8b.
\newblock {\em arXiv preprint arXiv:2406.07394}, 2024.

\bibitem{zhang2024negative}
Ruiqi Zhang, Licong Lin, Yu~Bai, and Song Mei.
\newblock Negative preference optimization: From catastrophic collapse to
  effective unlearning.
\newblock {\em arXiv preprint arXiv:2404.05868}, 2024.

\bibitem{zhao2023slic}
Yao Zhao, Rishabh Joshi, Tianqi Liu, Misha Khalman, Mohammad Saleh, and Peter~J
  Liu.
\newblock Slic-hf: Sequence likelihood calibration with human feedback.
\newblock {\em arXiv preprint arXiv:2305.10425}, 2023.

\bibitem{zhao2022calibrating}
Yao Zhao, Mikhail Khalman, Rishabh Joshi, Shashi Narayan, Mohammad Saleh, and
  Peter~J Liu.
\newblock Calibrating sequence likelihood improves conditional language
  generation.
\newblock In {\em The Eleventh International Conference on Learning
  Representations}, 2022.

\bibitem{zhao2023large}
Zirui Zhao, Wee~Sun Lee, and David Hsu.
\newblock Large language models as commonsense knowledge for large-scale task
  planning, 2023.

\bibitem{zhong2024dpo}
Han Zhong, Guhao Feng, Wei Xiong, Li~Zhao, Di~He, Jiang Bian, and Liwei Wang.
\newblock Dpo meets ppo: Reinforced token optimization for rlhf.
\newblock {\em arXiv preprint arXiv:2404.18922}, 2024.

\bibitem{zhou2024survey}
Zixuan Zhou, Xuefei Ning, Ke~Hong, Tianyu Fu, Jiaming Xu, Shiyao Li, Yuming
  Lou, Luning Wang, Zhihang Yuan, Xiuhong Li, et~al.
\newblock A survey on efficient inference for large language models.
\newblock {\em arXiv preprint arXiv:2404.14294}, 2024.

\end{thebibliography}

\clearpage
\onecolumn
\appendix
\section{Detailed Authors' Contributions}
\label{sec:contributions}
\textbf{Hanshi} co-lead the code infrastructure, led the implementation of the efficient inference engine and win-rate analysis, lead the final experiments in the paper and co-led the writing of the final paper \\
\textbf{Momin} co-lead the code infrastructure, led the preliminary experiments, and contributed to the writing of an early draft of the manuscript \\
\textbf{Ruiqi} provided several conceptual contributions to the work. He led the writing of the initial draft of the paper, and led the early statistical analysis to assess the feasibility of the project. \\
\textbf{Huitao} lead the theoretical part of the work \\
\textbf{Ming} provided useful feedback for the project during the weekly project meeting discussion \\
\textbf{Jiahao} contributed with a win-rate analysis during the rebuttal period  \\
\textbf{Mengdi} provided useful feedback and helped with accessing some of the compute infrastructure \\
\textbf{Peter} provided useful feedback, particularly regarding the correlation analysis in the early stage of the project and also co-suggested the iterative rejection scheme  \\
\textbf{Andrea} conceived the original idea of speculative rejection, advised the project, and co-led the final writing of the paper.

\section{Correlation between partial and final rewards}\label{appendix.correlation}
In this section, we present our observation that the partial and final rewards are positively correlative for the responses to a single prompt. We examine the distribution for the (empirical) Pearson correlation and Kendall's tau correlation coefficient for partial and final rewards for a single prompt. Mathematically, for $(\x_1,\x_2,...,\x_\n)$ and $(\y_1,\y_2,...,\y_\n),$ the two correlation are defined as
\begin{align*}
    R_{\mathsf{Pearson}} &:= \frac{\sum_{i=1}^\n (\x_i - \bar{\x}) (\y_i - \bar{\y})}{\sqrt{\sum_{i=1}^N (\x_i - \bar{\x})^2 \cdot \sum_{i=1}^N (\y_i - \bar{\y})^2}}, \\
    R_{\mathsf{Kendall}} &:= \frac{2}{\n (\n-1)} \sum_{i < j} \mathsf{sgn}(\x_i - \x_j) \cdot \mathsf{sgn}(\y_i - \y_j),
\end{align*}
where $\bar{\x} = \sum_{i=1}^N \x_i /N, \bar{\y} = \sum_{i=1}^N \y_i / N$ are their average, and $\mathsf{sgn}(\cdot)$ is the sign function.

\begin{figure}[H]
    \centering
    \includegraphics[width = 1\textwidth]{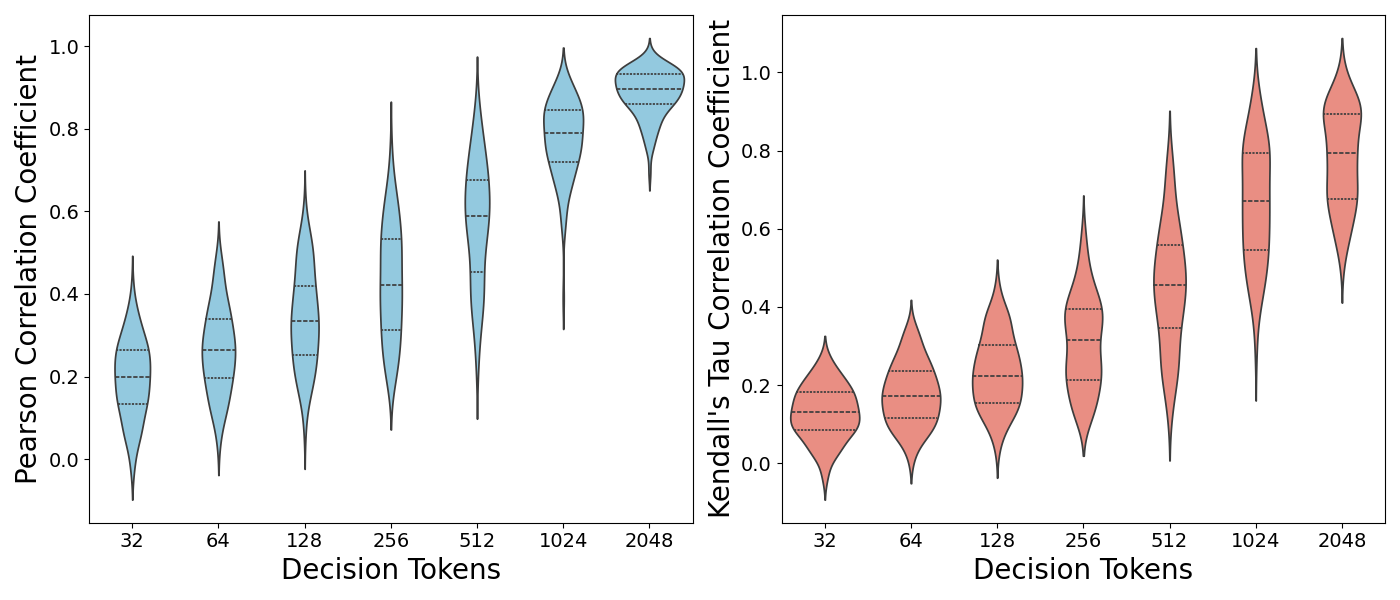}
    \caption{Pearson correlation (left) and Kendall's tau correlation coefficient (right) for the partial and final rewards. We randomly sample 100 prompts in the AlpacaFarm-Eval dataset. The responses are generated via Llama3-8b--Instruct and rewards are evaluated via Mistral-7B-RM.}
    \label{fig.correlation.appendix.}
\end{figure}

\newpage
\section*{NeurIPS Paper Checklist}

\begin{enumerate}

\item {\bf Claims}
    \item[] Question: Do the main claims made in the abstract and introduction accurately reflect the paper's contributions and scope?
    \item[] Answer: \answerYes{} % Replace by \answerYes{}, \answerNo{}, or \answerNA{}.
    \item[] Justification:  We believe our introduction and abstract are factually accurate in describing the contributions of the paper and of its result

\item {\bf Limitations}
    \item[] Question: Does the paper discuss the limitations of the work performed by the authors?
    \item[] Answer: \answerYes{} % Replace by \answerYes{}, \answerNo{}, or \answerNA{}.
    \item[] Justification: The limitations are presented in the final section of the main paper.

\item {\bf Theory Assumptions and Proofs}
    \item[] Question: For each theoretical result, does the paper provide the full set of assumptions and a complete (and correct) proof?
    \item[] Answer: \answerNA{} % Replace by \answerYes{}, \answerNo{}, or \answerNA{}.
    \item[] Justification: We do not include theoretical results in the paper.

    \item {\bf Experimental Result Reproducibility}
    \item[] Question: Does the paper fully disclose all the information needed to reproduce the main experimental results of the paper to the extent that it affects the main claims and/or conclusions of the paper (regardless of whether the code and data are provided or not)?
    \item[] Answer: \answerYes{} % Replace by \answerYes{}, \answerNo{}, or \answerNA{}.
    \item[] Justification: We provide all setup details to reproduce the experiment. In our supplementary zip file, we provide all code and evaluation datasets used, along with a README with instructions. We use public checkpoints for all draft and target models, public data for all evaluations.

\item {\bf Open access to data and code}
    \item[] Question: Does the paper provide open access to the data and code, with sufficient instructions to faithfully reproduce the main experimental results, as described in supplemental material?
    \item[] Answer: \answerYes{} % Replace by \answerYes{}, \answerNo{}, or \answerNA{}.
    \item[] Justification: The datasets that we use are freely available and hosted by reliable third parties. We provide a zip file with the code and data needed to reproduce our experiments, as well as a README with instructions.

\item {\bf Experimental Setting/Details}
    \item[] Question: Does the paper specify all the training and test details (e.g., data splits, hyperparameters, how they were chosen, type of optimizer, etc.) necessary to understand the results?
    \item[] Answer: \answerYes{} % Replace by \answerYes{}, \answerNo{}, or \answerNA{}.
    \item[] Justification: We only sample 100 prompts at random. There is no training involved.

\item {\bf Experiment Statistical Significance}
    \item[] Question: Does the paper report error bars suitably and correctly defined or other appropriate information about the statistical significance of the experiments?
    \item[] Answer: \answerNo{} % Replace by \answerYes{}, \answerNo{}, or \answerNA{}.
    \item[] Justification: The cost for producing the experiments precludes us from reporting error bars. However, notice that 
    each pair of llm and reward model produces a result that is averaged over 100 prompts. Since we report several such pairs, and the speedup is substantial for each single pair, we are confident that overall speedup is statistically significant.

\item {\bf Experiments Compute Resources}
    \item[] Question: For each experiment, does the paper provide sufficient information on the computer resources (type of compute workers, memory, time of execution) needed to reproduce the experiments?
    \item[] Answer: \answerYes{} % Replace by \answerYes{}, \answerNo{}, or \answerNA{}.
    \item[] Justification: This is discussed in the experiment section.
    
\item {\bf Code Of Ethics}
    \item[] Question: Does the research conducted in the paper conform, in every respect, with the NeurIPS Code of Ethics \url{https://neurips.cc/public/EthicsGuidelines}?
    \item[] Answer: \answerYes{} % Replace by \answerYes{}, \answerNo{}, or \answerNA{}.
    \item[] Justification: None to report.

\item {\bf Broader Impacts}
    \item[] Question: Does the paper discuss both potential positive societal impacts and negative societal impacts of the work performed?
    \item[] Answer: \answerNA{} % Replace by \answerYes{}, \answerNo{}, or \answerNA{}.
    \item[] Justification: This work aims at accelerating a known and existing method, and so it is not expected to have a direct societal impact.
    
\item {\bf Safeguards}
    \item[] Question: Does the paper describe safeguards that have been put in place for responsible release of data or models that have a high risk for misuse (e.g., pretrained language models, image generators, or scraped datasets)?
    \item[] Answer: \answerNA{} % Replace by \answerYes{}, \answerNo{}, or \answerNA{}.
    \item[] Justification: We do not release data or models.

\item {\bf Licenses for existing assets}
    \item[] Question: Are the creators or original owners of assets (e.g., code, data, models), used in the paper, properly credited and are the license and terms of use explicitly mentioned and properly respected?
    \item[] Answer: \answerYes{} % Replace by \answerYes{}, \answerNo{}, or \answerNA{}.
    \item[] Justification: We cite the papers that introduced the models and data used in our work.

\item {\bf New Assets}
    \item[] Question: Are new assets introduced in the paper well documented and is the documentation provided alongside the assets?
    \item[] Answer: \answerYes{} % Replace by \answerYes{}, \answerNo{}, or \answerNA{}.
    \item[] Justification:We include a README along with our code to reproduce our experiments.

\item {\bf Crowdsourcing and Research with Human Subjects}
    \item[] Question: For crowdsourcing experiments and research with human subjects, does the paper include the full text of instructions given to participants and screenshots, if applicable, as well as details about compensation (if any)? 
    \item[] Answer: \answerNA{} % Replace by \answerYes{}, \answerNo{}, or \answerNA{}.
    \item[] Justification: Our research does not involve humans.

\item {\bf Institutional Review Board (IRB) Approvals or Equivalent for Research with Human Subjects}
    \item[] Question: Does the paper describe potential risks incurred by study participants, whether such risks were disclosed to the subjects, and whether Institutional Review Board (IRB) approvals (or an equivalent approval/review based on the requirements of your country or institution) were obtained?
    \item[] Answer: \answerNA{} % Replace by \answerYes{}, \answerNo{}, or \answerNA{}.
    \item[] Justification: Our paper does not involve crowdsourcing.

\end{enumerate}

\end{document}